\pdfoutput=1

\documentclass[11pt]{article}
\usepackage[table]{xcolor}

\usepackage[]{acl}

\usepackage{times}
\usepackage{latexsym}
\usepackage{microtype}

\usepackage{fdsymbol}

\usepackage{microtype}
\usepackage{times}
\usepackage{latexsym}

\usepackage{soul}
\usepackage{url}
\usepackage[utf8]{inputenc}
\usepackage{graphicx}
\usepackage{booktabs}
\usepackage{colortbl}
\usepackage{multirow}
\usepackage{rotating}
\usepackage{listings}
\usepackage{gensymb}
\usepackage{enumitem}
\usepackage{lipsum}
\usepackage{placeins}   
\usepackage[linewidth=1pt]{mdframed}
\usepackage{xspace}

\usepackage[T1]{fontenc}

\usepackage[utf8]{inputenc}

\usepackage{tikz-dependency}
\usepackage{graphicx}
\usepackage{algorithmicx}
\usepackage[noend]{algpseudocode}
\usepackage{booktabs}
\usepackage[percent]{overpic}
\usepackage{subcaption}
\usepackage{siunitx}
\usepackage{wrapfig}
\usepackage{multirow}

\definecolor{bl}{RGB}{0,0,0}
\definecolor{gr}{RGB}{37,37,37}
\definecolor{sil}{RGB}{240,240,240}
\definecolor{gray}{RGB}{200,200,200}

\definecolor{green1}{HTML}{0b5400}
\definecolor{orange1}{HTML}{f3905c}
\definecolor{purple1}{HTML}{9258cc}
\definecolor{blue1}{HTML}{027db5}
\definecolor{pink1}{HTML}{ff7a7a}
\definecolor{darkgreen}{HTML}{005e19}
\definecolor{darkblue}{HTML}{240394}
\definecolor{darkred}{HTML}{C00000}
\definecolor{lightblue2}{HTML}{DEEBF7}
\definecolor{lightgreen2}{HTML}{E2F0D9}
\definecolor{lightgray2}{HTML}{767171}

\usepackage{listings}

\usetikzlibrary{arrows.meta}
\NewEnviron{elaboration}{
\par
\begin{tikzpicture}
\node[rectangle,minimum width=\textwidth,fill=none] (m) {\begin{minipage}{.98\textwidth}\BODY\end{minipage}};
\draw[rounded corners, dashed] (m.south west) rectangle (m.north east);
\end{tikzpicture}
}

\NewEnviron{arrowact}{
\par
\begin{tikzpicture}
\node[rectangle,minimum width=\textwidth,fill=white] (m) {\begin{minipage}{.98\textwidth}\BODY\end{minipage}};
\draw[->,>={LaTeX[scale=2]},draw opacity=0] (m.north) -- (m.south);
\end{tikzpicture}
}

\NewEnviron{elaborationskip}{
\par
\begin{tikzpicture}
\node[rectangle,minimum width=\textwidth,fill=gray] (m) {\begin{minipage}{.98\textwidth}\BODY\end{minipage}};
\path (m.south west) rectangle (m.north east);
\end{tikzpicture}
}

\NewEnviron{elaborationnar}{
\par
\begin{tikzpicture}
\node[rectangle,minimum width=\textwidth,fill=sil] (m) {\begin{minipage}{.98\textwidth}\BODY\end{minipage}};
\draw[rounded corners, dashed] (m.south west) rectangle (m.north east);
\end{tikzpicture}
}

\usepackage{enumitem}

\usepackage[export]{adjustbox}

\newcommand{\eg}{e.g.,\xspace}
\newcommand{\ie}{i.e.,\xspace}
\newcommand{\etc}{etc.\xspace}
\newcommand{\R}{\mathbb{R}}

\newcommand{\code}[1]{\fontdimen2\font=0.4em\texttt{#1}}
\newcommand{\cmd}[1]{\textcolor{darkgreen}{\textbf{\small{\code{#1}}}}}

\newcommand{\token}[1]{\textcolor{darkblue}{\textbf{\small{\code{#1}}}}}
\newcommand{\secondtoken}[1]{\textcolor{darkblue}{\textbf{\small{\code{#1}}}}}

\newcommand{\scienceworld}{\textsc{ScienceWorld}~}

\title{\textsc{ScienceWorld}: Is your Agent Smarter than a 5$^{\text{th}}$ Grader?}    

\author{Ruoyao Wang$^{*\ddagger}$, Peter Jansen$^{*\ddagger}$, Marc-Alexandre Côté$^{\clubsuit}$,
Prithviraj Ammanabrolu$^{\diamondsuit}$ \\
$^{\ddagger}$University of Arizona, Tucson, AZ  ~~~~~ $^{\clubsuit}$Microsoft Research Montréal \\
$^{\diamondsuit}$Allen Institute for AI, Seattle, WA \\
\texttt{\{ruoyaowang,pajansen\}@arizona.edu}\\ \texttt{macote@microsoft.com}, \texttt{raja@allenai.org}
}

\begin{document}
\maketitle
\begin{abstract}
We present \textsc{ScienceWorld}, a benchmark to test agents' scientific reasoning abilities in a new interactive text environment at the level of a standard elementary school science curriculum. 
Despite the transformer-based progress seen in question-answering and scientific text processing, we find that current models cannot reason about or explain learned science concepts in novel contexts.
For instance, models can easily answer {\em what} the conductivity of a known material is but struggle when asked {\em how} they would conduct an experiment in a grounded environment to find the conductivity of an unknown material.
This begs the question of whether current models are simply retrieving answers by way of seeing a large number of similar examples or if they have learned to reason about concepts in a reusable manner.
We hypothesize that agents need to be grounded in interactive environments to achieve such reasoning capabilities.
Our experiments provide empirical evidence supporting this hypothesis---showing that a 1.5 million parameter agent trained interactively for 100k steps outperforms a 11 billion parameter model statically trained for scientific question-answering and reasoning from millions of expert demonstrations.
\footnote{Website: \href{https://sciworld.apps.allenai.org}{https://sciworld.apps.allenai.org}}\footnote{Code: \href{https://github.com/allenai/ScienceWorld}{https://github.com/allenai/ScienceWorld}}

\end{abstract}

\section{Introduction}
\begin{table}[t!]
\begin{center}
\scriptsize
\setlength{\tabcolsep}{3pt}
\begin{tabular}{p{0.98\linewidth}} 
\toprule
    \textbf{\textsc{ScienceWorld} Task 3 (Test Electrical Conductivity Subtask)} \\
\midrule
\includegraphics[width=1.0\columnwidth]{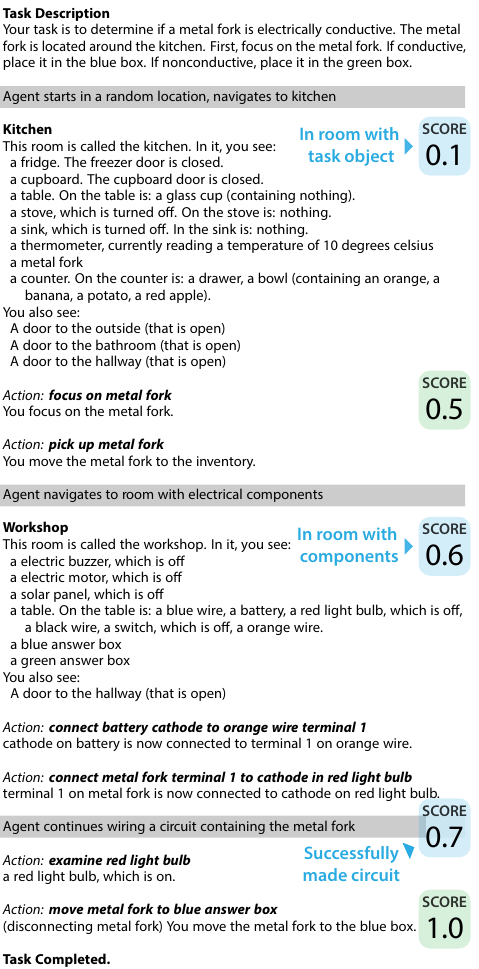}\\
\bottomrule
\end{tabular}
\caption{\footnotesize Transcript for one of the 30 tasks (testing an object's electrical conductivity, here a \textit{metal fork}) in \textsc{ScienceWorld}. 
The agent finds the \textit{metal fork}, use it to build an electrical circuit with a \textit{red light bulb}, and observe if the light turns on. Note the transcript has been simplified to fit in available space, and large sections have been omitted (greyed sections).
\label{tab:scienceworld-example}}
\end{center}
\vspace{-4mm}
\end{table}

Question answering (QA) has seen rapid progress recently.
Standardized elementary and middle school science exams have served as a challenge task for QA \cite{clark2018think}, as these questions require combining science-domain knowledge with world knowledge in complex reasoning procedures to solve. As large language models have toppled these benchmarks \cite{clark2020f,khashabi-etal-2020-unifiedqa,xu-etal-2021-exploiting-reasoning}, the focus has shifted away from simply answering questions toward producing human-readable explanations for a model's answers \cite{jansen-etal-2018-worldtree,yang-etal-2018-hotpotqa,xie-etal-2020-worldtree,valentino-etal-2021-unification,xu-etal-2021-dynamic,lamm-etal-2021-qed,aggarwal-etal-2021-explanations}.

%
%

While language models are able to produce compelling answers \cite{zoph2022designing} or explanations \cite{jansen-etal-2021-challenges} to science questions, are they simply retrieving (or shallowly assembling) these answers, or can they understand and use the knowledge they output in a meaningful way? Also, how can we evaluate if a model's explanation is correct?



In this work we explore these two questions by reframing science exam question answering into an interactive task where agents must complete elementary science experiments in a simulated text-based environment called \textsc{ScienceWorld} (see Table~\ref{tab:scienceworld-example}).  Instead of simply answering a question \textit{(\eg Q: ``What will happen to an ice cube when placed on a stove?'', A: ``it will melt''}), the agent must 
demonstrate its capacity to combine declarative scientific knowledge with the procedural knowledge required to correctly complete the experiment in the virtual environment.
Similarly, the sequence of virtual actions an agent performs can serve as a form of procedural \textit{(``how'')} explanation to the question, that can be directly evaluated in the virtual environment for correctness (\eg whether the actions led to the ice cube melting). 

\paragraph{The contributions of this work are:}
\begin{enumerate}
    \vspace{-1em}
    \item We construct \textsc{ScienceWorld}, a complex interactive text environment, with simulation engines for thermodynamics, electrical circuits, chemistry reactions, and biological processes. 
    \item We implement 30 benchmark tasks across 10 topics spanning the elementary science curriculum, including changes of state of matter and the role of pollinators when growing fruit.
    \item We evaluate 5 state-of-the-art reinforcement learning and language model agents on this benchmark, empirically showing that they
    perform poorly on tasks (\eg melting ice) that 5$^{\text{th}}$ grade students can perform with ease. 
\end{enumerate}

\begin{figure}[]
\centering
\includegraphics[width=0.98\columnwidth]{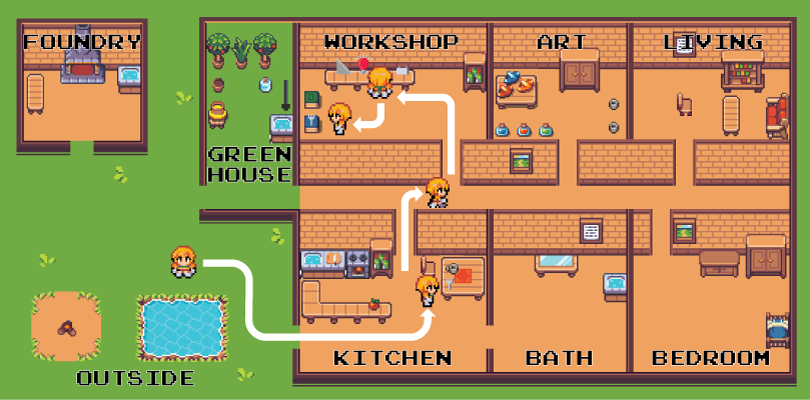}
\caption{\footnotesize A graphical representation of an agent performing the Electrical Conductivity Subtask in \textsc{ScienceWorld}. }
\label{fig:overview}
\end{figure}

\section{Related Work}
{\flushleft \textbf{Science-domain Inference:}} 
Standardized science exams are a challenging task for question answering due to their diverse knowledge and inference requirements \cite{Clark2013ASO,jansen-etal-2016-whats,boratko-etal-2018-systematic,clark2018think}.
Top performing models can answer
more than 90\% of multiple-choice questions correctly \cite{clark2020f}, typically through the use of large language models \cite{khashabi-etal-2020-unifiedqa,zoph2022designing}.

A number of corpora of structured and semi-structured science exam explanations exist for training the explanation-generation task \cite{jansen-etal-2018-worldtree,xie-etal-2020-worldtree,khot2020qasc,dalvi-etal-2021-explaining}. 
Evaluating explanations is challenging, and typically done by comparing generated explanations to a single gold explanation. This has been shown to substantially underestimate explanation generation performance by up to~40\% \cite{jansen-etal-2021-challenges}. 
While others have worked to mitigate this by generating multiple alternate explanations for each question, this is expensive, and has only been demonstrated for adding one or two additional explanations per question \cite{inoue-etal-2020-r4c,jhamtani-clark-2020-learning}. 
Here, we propose a partial solution to this evaluation problem by treating the action sequences of agents as structured \textit{manner} explanations for \textit{how} to solve a task.  These action sequences can be directly run in the \textsc{ScienceWorld} simulator to automatically determine their correctness (\ie whether they accomplish the task), independent of any variations in their solution methods.  For example, whether an agent melts ice by using a stove, building a campfire, or leaving the ice out on a kitchen counter, the end result is the same and directly measurable through the formal semantics of the simulator. 


{\flushleft \textbf{Environments:}} Interactive text environments are becoming a vehicle for research in natural language processing (see \citet{jansen2021systematic} for a review), primarily because of their reduced development costs relative to 3D environments, combined with their ability to easily implement high-level tasks with large action spaces.  In spite of these benefits, implementation costs can still be substantial for large complex environments, and text agents are frequently evaluated on a small set of existing interactive fiction games such as Zork \cite{lebling1979zork} using an unified interface like Jericho \cite{hausknecht2020interactive}.
A few purpose-built environments provide simple tasks for studying text agents, typically on procedurally generated pick-and-place or object-combination tasks \cite[\eg cooking,][]{yin2019learn}.
Kitchen Cleanup \cite{murugesan2020enhancing} and TextWorld Common Sense \cite{murugesan2020text} require agents to tidy up one or more rooms in a house environment by putting objects in their typical locations (\eg a \textit{hat} should be placed on the \textit{hat rack}), testing an agent's declarative knowledge of common object locations with the procedural knowledge required for this pick-and-place task.  The closest existing interactive text environment to \textsc{ScienceWorld} is TextLabs \cite{tamari-etal-2021-process} which simulates chemistry wet-lab protocols with actions such as \textit{pipetting} and \textit{centrifuging}. Compared to these existing environments, \textsc{ScienceWorld} is generally a larger and more dynamic environment, populated with more complex objects with greater depth of physical simulation.  This simulation depth enables more complex tasks associated with elementary science (\eg thermodynamics, electrical circuits, etc.) to be tested, and a greater variety of solutions.  

{\flushleft \textbf{Simulators:}} Nearly all text-based world simulations are currently implemented as Z-machine games \cite{learningZIL1989,nelson2014zmachine}, frequently through higher-level application-specific languages \cite[such as Inform7,][]{nelson2006natural} that compile to Z-machine code.  TextWorld \cite{cote2018textworld} creates environments using linear-logic statements that specify action preconditions and postconditions \cite{martens2015ceptre} and generate Inform7 code. 
Existing tooling is designed for simpler simulations than \textsc{ScienceWorld}, with primarily agent-centered state changes that make modeling autonomous physical processes (\eg thermodynamics) difficult. As such, in this work we build a novel simulator to model physical processes in text environments. 


{\flushleft \textbf{Agents:}} A variety of agent models have been proposed for reasoning in interactive text environments.  Most approaches frame reasoning as a partially-observable Markov decision process (POMDP), and model inference using reinforcement learning (RL).  This includes RL-based models that learn a policies to pick relevant actions from lists of candidate actions \cite{he-etal-2016-deep-reinforcement}, or models that mix RL with knowledge graphs \cite{ammanabrolu2020graph} or language models \cite{yao-etal-2020-keep} to aid in next-action selection.  Action selection has also been modelled using case-based reasoning \cite{atzeni2021case}, or directly as a sequence-prediction imitation learning task, using language models trained on gold action sequences to predict the next action given the current state \cite{torabi2018,ammanabrolu-etal-2021-motivate,ammanabrolu2022dialogue}.  
In general, agent performance on solving interactive fictions is still modest, with only easier games close to completion \cite{jansen2021systematic}. 

In this work, we benchmark state-of-the-art agents on \textsc{ScienceWorld} as well as introduce novel agents. We empirically show that these elementary science tasks are difficult for current agents, and also that smaller and simpler agents can outperform billion-scale parameter language models trained on gold sequences, highlighting the difficulty of this task for transformer-based models. 







\section{\textsc{ScienceWorld}}
\textsc{ScienceWorld} is a simulation of the world abstracted through a complex interactive text environment in English with many objects, actions, and simulation engines. The framework consists of 40k lines of \textsc{Scala} (speed) with a \textsc{Python} interface.

The \textsc{ScienceWorld} environment contains 10 interconnected locations (see Figure~\ref{fig:overview}), populated with up to 200 types of objects, including devices, instruments, plants/animals, electrical components, substances, containers, and common environment objects such as furniture, books, and paintings. The \textsc{ScienceWorld} action space contains 25 high-level actions, including both science-domain actions (\eg using \token{thermometer}) and common actions (\eg moving, opening containers, picking up items), with approximately 200k possible action-object combinations per step (though only a limited subset of these will be meaningful).  See Appendix~\ref{sec:appendix-scienceworld-description} for details about \textsc{ScienceWorld}, including the object model, actions, and input parser.

\subsection{Simulation Engines}
\label{sec:Simulation-Engines}
\textsc{ScienceWorld} supports actions commonly found in interactive text environments -- for example, objects can be moved or examined, foods can be eaten, and books can be read.  In addition, the environment contains a number of elementary science-domain specific processes that either occur automatically (\eg thermodynamics) or are coupled to actions (\eg devices, mixing chemicals). 
Those simulation engines\footnote{
For tractability, simulation engines are implemented with fidelity at the level of elementary science. Thermal transfer uses a simplified equation, biological changes happen in stages rather than gradually, only simple series circuits are simulated (no resistance, inductance, or any advanced topics), etc.} 
are:
 
{\flushleft \textbf{Thermodynamics:}} All objects have temperatures and other thermal properties based on their materials.  All objects within a container are considered in thermal contact with each other, and transfer heat energy using a simplified conductive heat model.  The proportion of heat transferred between objects at each step is mediated by the object's \textit{thermal conduction coefficient}, allowing thermal conductors (like metal pots) and insulators (like ceramics) to be modelled.  Every material has phase transition points \textit{(\ie melting point, boiling point)} and \textit{combustion points} populated based on the best-known or approximate physical values for those materials.  Objects that move past these thresholds will change state of matter (\eg from a solid to a liquid), or begin a combustion process that ultimately ends in the object turning to ash unless its fire is put out.  Convective heat transfer is also modelled in the form of heat sources (\eg \token{oven}, \token{stove}) and heat sinks (\eg \token{fridge}, \token{freezer}) that transfer heat energy to or from objects.  Rooms also transfer ambient heat energy to/from the objects they contain.

{\flushleft \textbf{Electricity:}} The simulator models simple series electrical circuits, where electrically-powered devices (\eg \token{light bulb}, \token{motor}) can be powered by being connected to electrical energy sources (\eg \token{battery}, \token{solar panel}) through electrical conductors (nominally, \token{wires}).  Polarized and unpolarized components are modelled, with each object having exactly two terminals (anode and cathode for polarized; terminals 1 and 2 for unpolarized).  Connection happens through explicit terminal-to-terminal connection actions (\eg \cmd{connect \token{battery anode} to \token{blue wire terminal 1}}).  Every non-electrical object in \textsc{ScienceWorld} has virtual unpolarized terminals, allowing circuits to be build with valid electrical conductors (\eg using a \token{metal fork} in place of a \token{wire}), and for the agent to build circuits that test conductivity by (for example) observing if a light bulb illuminates when a plastic versus metal fork is used in the circuit.

{\flushleft \textbf{Devices:}} Many objects are also considered devices, that can be activated or deactivated by the agent (\eg \token{stove}, \token{electrical switch}), or may have environment-specific conditions to being activated (\eg a \token{light bulb} will only activate if it is properly electrically connected; a \token{solar panel} will only produce power if it is outside). Objects can also be used with other objects in specific contexts (\eg a \token{thermometer}, to measure an object's temperature; a \token{shovel}, to dig soil from the ground).  

{\flushleft \textbf{Chemistry:}} A subset of specific chemical reactions are modelled, where mixing a set of substances together in a container will produce a resultant substance (\eg \token{salt} and \token{water} mix to produce \token{salt water}).  Common chemical reactions taught in elementary science (\eg water reactions, rust, food reactions, paint mixing) are modelled.

{\flushleft \textbf{Life Stages:}} Living things (plants and animals) progress through life stages (\eg seed, seedling, juvenile plant, adult plant, reproducing plant, dead plant).  Progression through life stages happens over time by continuing to meet the needs of that living thing (\eg \token{water}, \token{soil}).  If the needs are not met (\eg a plant is not watered, is removed from soil, or becomes too hot), then it dies.

{\flushleft \textbf{Reproduction and Genetics:}} Living things can have genes that express traits (\eg flower colour, seed shape, leaf size).  Genes are inherited from the alleles of both parents, and genotype is determined at the time of reproduction using a Punnett square.  Phenotype (expressed, visible traits) are determined based on which genes are dominant versus recessive.  Currently, traits are only populated for selected plants to reproduce Mendelian-genetics experiments.  Plants reproduce by exchanging pollen (containing their genes) between flowers, typically by a pollinator (such as a \token{bee}).  Pollinated flowers eventually wilt and turn into fruits containing seeds of genetic descendants. 

{\flushleft \textbf{Friction (Inclined Plane):}} Forces are a significant part of an elementary science curriculum, but difficult to incorporate without 2D or 3D simulation.  \textsc{ScienceWorld} models the forces of gravity and friction in the specific context of 1-dimensional inclined plane experiments.  Objects placed at the top of an inclined plane will slide down the plane at a speed proportional to the plane's angle, and the friction coefficient of its surface material.  The position is described to the agent (\eg \textit{``an inclined plant with a block 60\% of the way down the plane''}), allowing experiments to determine either the relative angle or friction coefficients of different inclined planes based on the speed the object moves down a given plane. 

{\flushleft \textbf{Containers:}} Containers can be always open (\eg a \token{metal pot}) or closeable (\eg a \token{cupboard}).  Objects contained inside containers are not visible until the container is open.  Some effects spread beyond a container -- for example, a wooden cupboard with a hot object inside may combust, causing other objects in the kitchen to also increase temperature.

\section{Experiments}

To understand how contemporary approaches to text agents perform at \textsc{ScienceWorld} tasks, we benchmark a selection of recent architectures.  

{\flushleft\textbf{Tasks.}} To support our goal of generating a diverse set of tasks, we identified a candidate set of 10 broad science exam topics from the list of 400 fine-grained science curriculum topics of \citet{xu-etal-2020-multi}.  
Topics were chosen that would be amenable to text-based simulation, and that did not have critical fine-grained spatial reasoning requirements, and include: changes of state, temperature measurement, electrical circuits, friction, object classification, chemical mixtures, plants and pollinators, life spans, life stages, and Mendelian genetics.  
Each topic was further divided into between 2 and 4 specific tasks for agents to perform, producing a total of 30 science-domain tasks.
These topics and tasks are described in Appendix ~\ref{sec:specifictasks}.

To prevent overfitting and encourage generalization, each subtask contains between 10 and 1400 parametric variations (with 7200 total variations across all 30 subtasks).  Variations change critical task objects (\eg the specific substance to be melted), the agent's starting location in the environment, as well as randomly vary the contents of the environment itself (\eg whether the living room contains a bookshelf, or a painting, or both). 

{\flushleft\textbf{Train, Development, Test sets:}} For a given subtask, variations are split into 50\% training, 25\% development, and 25\% test sets.  Variations are sorted such that critical unseen variations (\eg substances, animals, or plants unseen during training) are found in development and test sets. 

{\flushleft\textbf{Goals and Rewards.}} To reduce reward sparsity, each task includes between 2 and 15 optional subgoals (such as \textit{turning on the stove}, or \textit{the substance increasing in temperature by 10C}) that help nudge agents in the direction of canonical solutions, if desired.  Meeting required and optional subgoals increases the agent's score on a given subtask.  Scores for all tasks are normalized to between 0 and 1. 

{\flushleft\textbf{Oracle Agents}} To support imitation learning, we provide gold trajectories from 30 hand-coded oracles on all subtasks and variations.  For tractability these solutions represent canonical solution methods (\eg using a stove to boil water), rather than all possible solution methods that lead to the goal state (\eg building a campfire to boil water). 

{\flushleft\textbf{Learning Agents.}}
An interactive text environment can be cast as a partially observable Markov decision process (POMDP) defined by $\langle S, T, A, R, O, \Omega, \gamma \rangle$ representing the set of possible states ($S$), conditional transition probabilities between states ($T$), available text commands ($A$), reward function ($R:S\times A\rightarrow\R$), set of possible text observations ($O$), observation conditional probabilities ($\Omega:S\rightarrow O$), and discount factor ($\gamma\in[0,1]$). The goal for a learning agent is then to learn a policy $\pi_\theta(o)\rightarrow{a}$ that chooses or generates a text command $a\in A$ given the text observation $o\in\Omega(s)$ of state $s\in S$ that maximizes the expected discounted sum of rewards $\mathop{\mathbb{E}}[\sum_t\gamma^tR(s_t,a_t)]$. In \textsc{ScienceWorld}, the agent is also provided with a task description $d$.

To provide a fair comparison, all models were run using identical experiment configurations when possible.
Reinforcement learning models were run with identical training regiments (8 environment threads at 100k steps per thread).  Training episodes reset after meeting an end state (success or failure), or after reaching a maximum number of steps (we used 100 in all experiments).
Additional model details are provided in Appendix~\ref{sec:appendixDetails}. 

{\flushleft \textbf{Random Baseline}:}  At each time step $t$, this baseline randomly chooses an action $a$ from the set of valid actions $A_t$ obtained from the simulator. 

{\flushleft \textbf{DRRN} \cite{he-etal-2016-deep-reinforcement}:} The Deep Reinforcement Relevance Network (DRRN) learns separate representations of the observation space and action space of an environment, then trains a policy that selects from $A_t$ the action that is the most relevant given the current text observation $o_t$ (which also includes the description of the current room $o_t^\textrm{look}$ and the current agent inventory $o_t^\textrm{inv}$)
and that would lead to an increased reward.  
The DRRN is a strong baseline with near state-of-the-art performance on many medium-to-hard interactive text environments~\citep{hausknecht2020interactive}. 

{\flushleft \textbf{KG-A2C} \cite{ammanabrolu2020graph}:} This model represents the state space with a knowledge graph built dynamically from the text observations $o_t$ using OpenIE triples \cite{angeli-etal-2015-leveraging} such as \textit{(\token{glass bottle}, \cmd{contains}, \token{water})}, while the action space is represented using action templates with placeholders (\eg \cmd{open \token{OBJ}}) obtained from \textsc{ScienceWorld}.  
The model learns a policy that selects relevant action templates then populates them with objects from the knowledge graph. 


{\flushleft \textbf{CALM (GPT2)} \cite{yao-etal-2020-keep}:} We collect transcripts of expert demonstrations for the train variations of the tasks using the oracle agents, then use them to fine-tune a pre-trained language model (GPT-2, \citet{radford2019language}).
At runtime, the language model is provided with the current observation $o_t$, last observation $o_{t-1}$, and last action $a_{t-1}$, then generates a shortlist of 30 possible actions to take.  This shortlist serves as input to an RL model similar to the DRRN, which re-ranks the actions and chooses the next action to perform.

{\flushleft \textbf{Behavior Cloning}~\citep{torabi2018}:}
We follow the methodology of \citet{ammanabrolu-etal-2021-motivate} in adapting the popular imitation learning method of behavior cloning from observations to text agents. We used the same transcripts of demonstrations as the CALM (GPT2) agent to extract 211,092 training examples with $(d, o_{t-1}, a_{t-1}, o_t)$ as inputs and $a_t$ as targets. We fine-tune a transformer-based text-to-text model with a T5 architecture~\citep{2020t5} initialized with the weights of a Macaw~\citep{Tafjord2021Macaw} model designed to answer science questions.

At test time, the agent performs zero-shot inference online in the simulator by generating a fixed number of actions with beam search on the unseen test variations for each task.
Despite training on a large number of demonstrations, the generated actions are often invalid or not useful---resulting in zero scores. Thus, we treat the beam search's output as a ranked-list and run the highest-ranked action appearing in $A_t$,
similar to the language-model-to-valid-action aligner of \citet{huang2022language}.  

{\flushleft \textbf{Text Decision Transformer}:}
Inspired by the Decision Transformer~\citep{chen2021decisiontransformer}, we create a novel text game agent that models the entire POMDP trajectory as a sequence and has the ability to potentially predict actions that maximize future long term expected reward. We again used the same transcripts of demonstrations as the two previous agents to extract 224,902 training examples with $(d, o_{t-1}, \hat{R}_{t-1}, a_{t-1}, o_t, \hat{R}_t)$ as inputs and $a_t$ as targets.
Here $\hat{R}$ is the returns-to-go (\ie sum of future rewards) $\hat{R}=\sum_{t'=t}^T r_{t'}$ where $r_{t'}$ is the future reward obtained by the expert at step $t'$---enabling models to predict actions that maximize future expected rewards.
The architecture, pre-training, parameter sizes, and test inference are otherwise similar to the behavior cloning agent. 

Both the Behavior Cloning and Text Decision Transformer agents learn to perform \scienceworld tasks offline from demonstrations once pre-trained for scientific QA.
They use the prevailing paradigm for achieving state-of-the-art on many language benchmarks (\eg QA~\citep{khashabi-etal-2020-unifiedqa,Tafjord2021Macaw}, language understanding~\citep{2020t5,gpt3}).

\begin{table*}[!t]
\centering
\scriptsize
\vspace{8mm}
\begin{tabular}{lllcccccccc}
\textbf{\#} & \textbf{Topic} & \textbf{Task} &
Random-Valid & DRRN & KG-A2C & CALM & BC & TDT \\
\toprule
     \multicolumn{3}{l}{\textbf{Rely on Test Time Valid Action Detection Aid}}  &                   $\checkmark$                 & $\checkmark$      &     &   &   $\checkmark$  &  
     $\checkmark$  \\

\toprule
\rowcolor[HTML]{D3D3D3} 
1-1  & Matter         & Changes of State (Boiling)                  & 0.00 & 0.03 & 0.00 & 0.00 & 0.00 & 0.00 \\ 
\rowcolor[HTML]{D3D3D3} 
1-2  & Matter         & Changes of State (Melting)                  & 0.00 & 0.04 & 0.00 & 0.00 & 0.00 & 0.01 \\ 
\rowcolor[HTML]{D3D3D3} 
1-3  & Matter         & Changes of State (Freezing)                 & 0.00 & 0.01 & 0.04 & 0.00 & 0.01 & 0.00 \\ 
\rowcolor[HTML]{D3D3D3} 
1-4  & Matter         & Changes of State (Any)                      & 0.00 & 0.03 & 0.00 & 0.00 & 0.00 & 0.00 \\ \hline  
2-1  & Measurement    & Use Thermometer                             & 0.00 & 0.10 & 0.06 & 0.01 & 0.04 & 0.04 \\ 
2-2  & Measurement    & Measuring Boiling Point (known)             & 0.00 & 0.08 & 0.11 & 0.01 & 0.01 & 0.02 \\ 
2-3  & Measurement    & Measuring Boiling Point (unknown)           & 0.00 & 0.06 & 0.04 & 0.01 & 0.01 & 0.02 \\ \hline 
\rowcolor[HTML]{D3D3D3} 
3-1  & Electricity    & Create a circuit                            & 0.01 & 0.13 & 0.07 & 0.05 & 0.03 & 0.07 \\ 
\rowcolor[HTML]{D3D3D3} 
3-2  & Electricity    & Renewable vs Non-renewable Energy           & 0.01 & 0.10 & 0.04 & 0.07 & 0.02 & 0.05 \\ 
\rowcolor[HTML]{D3D3D3} 
3-3  & Electricity    & Test Conductivity (known)                   & 0.01 & 0.07 & 0.04 & 0.02 & 0.05 & 0.05 \\ 
\rowcolor[HTML]{D3D3D3} 
3-4  & Electricity    & Test Conductivity (unknown)                 & 0.00 & 0.20 & 0.04 & 0.02 & 0.04 & 0.05 \\ \hline 
4-1  & Classification & Find a living thing                         & 0.03 & 0.26 & 0.18 & 0.10 & 0.29 & 0.16 \\ 
4-2  & Classification & Find a non-living thing                     & 0.63 & 0.56 & 0.44 & 0.54 & 0.19 & 0.17 \\ 
4-3  & Classification & Find a plant                                & 0.01 & 0.19 & 0.16 & 0.10 & 0.17 & 0.19 \\ 
4-4  & Classification & Find an animal                              & 0.01 & 0.19 & 0.15 & 0.08 & 0.21 & 0.19 \\ \hline 
\rowcolor[HTML]{D3D3D3} 
5-1  & Biology        & Grow a plant                                & 0.07 & 0.09 & 0.06 & 0.02 & 0.08 & 0.03 \\ 
\rowcolor[HTML]{D3D3D3} 
5-2  & Biology        & Grow a fruit                                & 0.02 & 0.16 & 0.11 & 0.04 & 0.03 & 0.05 \\ \hline 
6-1  & Chemistry      & Mixing (generic)                            & 0.01 & 0.20 & 0.17 & 0.03 & 0.06 & 0.10 \\ 
6-2  & Chemistry      & Mixing paints (secondary colours)           & 0.01 & 0.29 & 0.19 & 0.06 & 0.16 & 0.20 \\ 
6-3  & Chemistry      & Mixing paints (tertiary colours)            & 0.00 & 0.11 & 0.04 & 0.03 & 0.05 & 0.07 \\ \hline 
\rowcolor[HTML]{D3D3D3} 
7-1  & Biology        & Identify longest-lived animal               & 0.02 & 0.48 & 0.43 & 0.06 & 0.26 & 0.20 \\ 
\rowcolor[HTML]{D3D3D3} 
7-2  & Biology        & Identify shortest-lived animal              & 0.03 & 0.47 & 0.32 & 0.10 & 0.14 & 0.16 \\ 
\rowcolor[HTML]{D3D3D3} 
7-3  & Biology        & Identify longest-then-shortest-lived animal & 0.01 & 0.31 & 0.23 & 0.04 & 0.02 & 0.20 \\ \hline 
8-1  & Biology        & Identify life stages (plant)                & 0.00 & 0.09 & 0.05 & 0.04 & 0.04 & 0.02 \\ 
8-2  & Biology        & Identify life stages (animal)               & 0.00 & 0.10 & 0.10 & 0.00 & 0.02 & 0.07 \\ \hline 
\rowcolor[HTML]{D3D3D3} 
9-1  & Forces         & Inclined Planes (determine angle)           & 0.01 & 0.13 & 0.04 & 0.00 & 0.05 & 0.04 \\ 
\rowcolor[HTML]{D3D3D3} 
9-2  & Forces         & Friction (known surfaces)                   & 0.00 & 0.13 & 0.04 & 0.03 & 0.05 & 0.04 \\ 
\rowcolor[HTML]{D3D3D3} 
9-3  & Forces         & Friction (unknown surfaces)                 & 0.01 & 0.13 & 0.04 & 0.02 & 0.04 & 0.04 \\ \hline 
10-1 & Biology        & Mendelian Genetics (known plants)           & 0.01 & 0.19 & 0.11 & 0.02 & 0.06 & 0.06 \\ 
10-2 & Biology        & Mendelian Genetics (unknown plants)         & 0.01 & 0.17 & 0.11 & 0.02 & 0.13 & 0.05 \\ 
\midrule
     \multicolumn{2}{l}{\textbf{Average}}  &                                    & {0.03}      &   \textbf{0.17}  &       {0.11}  &   {0.05}  &   {0.08}  &   {0.08}\\
        \multicolumn{2}{l}{\textbf{Param. Count $\times10^6$}}  &                                   &  --    &  \textbf{1.5}   &    5.5     &  131\textsuperscript{*}   &   11,000  & 11,000  \\
\bottomrule
\end{tabular}
\caption{\footnotesize Zero-shot performance of the agents on test variations of across all tasks. All online RL-trained agent performances are averaged over 5 independent random seeds.  Results across seeds tend to have low variance, with 80\% of standard deviations below 0.05, and 95\% of standard deviations below 0.10. Performance for RL agents is averaged over the last 10\% of evaluation episodes, while T5 performance represents average task score across all test variations of a task.\footnotesize $^{*}$ signifies that the value of 131M parameters includes the number of the parameters of the pre-trained GPT-2 action generator model. Only 6.9 million policy parameters are updated in RL training. }
\label{tab:agent-performance} 
\end{table*}

%
%
\begin{figure*}[t!]
\vspace{6mm}
\centering
\begin{subfigure}{.24\textwidth}
  \includegraphics[width=\textwidth]{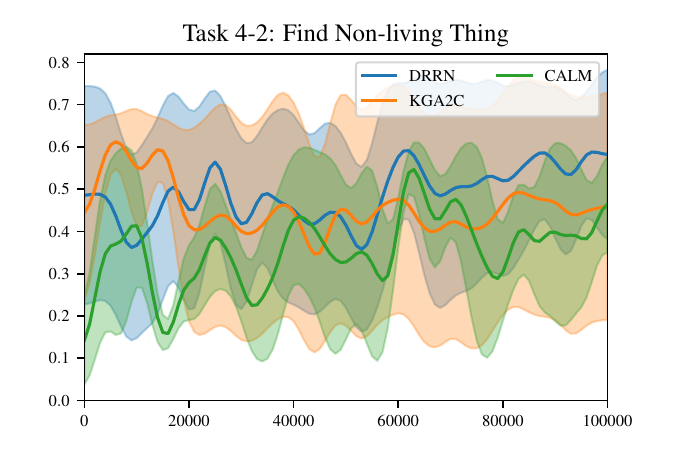}
\end{subfigure} 
\begin{subfigure}{.24\textwidth}
  \includegraphics[width=\textwidth]{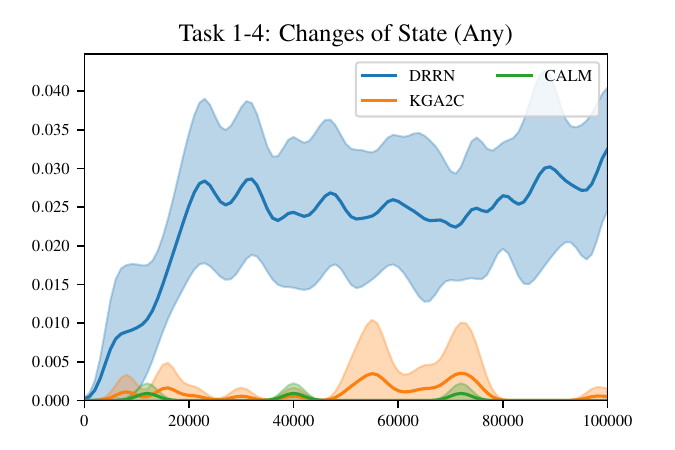}
\end{subfigure}
\begin{subfigure}{.24\textwidth}
  \includegraphics[width=\textwidth]{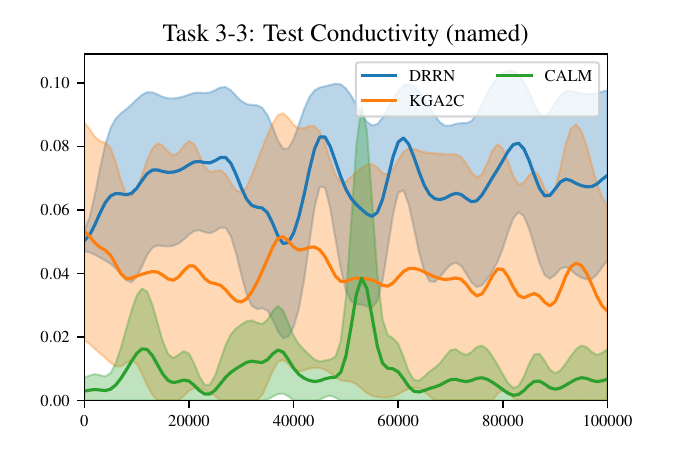}
\end{subfigure} 
\begin{subfigure}{.24\textwidth}
  \includegraphics[width=\textwidth]{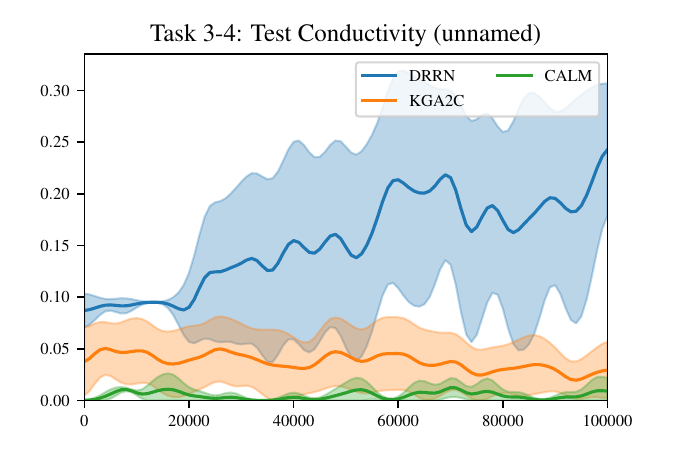}
\end{subfigure}
\caption{\footnotesize Episode reward curves for the DRRN, KGA2C, and CALM models on the unseen test set as a function of the number of training environment interactions. \textit{(left)} An example of an easier task, where the agent must pick up any non-living thing, and place it in a specific box in the environment.  \textit{(center-left)} An example challenge task, where the agent must perform any change of state (melt, boil, freeze) on a specific substance.  \textit{(right)} Performance on two variations of a task where the agent must determine whether a specific substance is electrically conductive or not.  In one variation \textit{(center-right)}, the substance is named (e.g. metal fork), while in the other variation \textit{(right)} the substance is randomly generated (e.g. unknown substance B).}
\label{fig:rlcurves1}
\vspace{6mm}
\end{figure*}

{\flushleft\textbf{Results.}}
Performance for all agents across each \textsc{ScienceWorld} task is shown in Table~\ref{tab:agent-performance}.  Overall, these tasks are challenging for current models, with the best model (DRRN) achieving an average score of 0.17 across all 30 subtasks.  Models that rely on the valid action detection aid 
generally perform better than those that learn to generate \textit{valid} actions in addition to learning what actions to pick to increase task performance.  All models relying on large language model for action selection (CALM, BC-T5, TDT-T5) generally achieve low performance as they tend to generate few valid actions in their candidate action lists.

Figure~\ref{fig:rlcurves1} shows episode reward curves for four selected tasks (with reward curves for all tasks included in Appendix~\ref{sec:appendixDetails}).  These four tasks include the current best-performing task \textit{(Task 4-2: Find a non-living thing)}, which requires an agent to focus on any non-living thing in the environment, pick it up, and place it in a specific container (typically in a different location than the agent's starting location).  Most RL models quickly solve the majority of this task, but struggle with picking up and moving the object to the final container. In contrast, other more open-ended tasks (such as \textit{Task 1-4}, which requires agents to perform any state-of-matter change to a specific substance) are performed poorly by all models.  Finally, \textsc{ScienceWorld} includes pairs of identical tasks where one can be solved by retrieving some critical component of the answer, while the other requires conducting the experimental procedure successfully.  For example, in \textit{Task 3-3}, an agent could look up that a \token{metal fork} is an electrical conductor and solve the task with comparatively fewer steps then in its paired \textit{Task 3-4}, where the substance name is randomly generated (\eg \token{unknown substance B}) and the experiment must be completed to get the answer.  We do not yet observe this behavior with the agents under examination. They generally struggle with commonsense level tasks (\eg navigation) and are unable to reach a point where the language models (either GPT-2 in CALM, or T5 initialized with Macaw in BC and TDT) are able to leverage their internal knowledge to solve these tasks through retrieval.




\section{Discussion}

{\flushleft\textbf{Elementary science tasks are challenging for text agents.}} With top-performing agents reaching normalized average scores of 0.17 across tasks, performance on \textsc{ScienceWorld} is comparable to the current best-performing agents on medium-difficulty interactive fiction games such as Zork \cite{ammanabrolu2020avoid,yao2021reading}.  Much as in interactive fiction games, examining agent trajectories reveals that while agents appear to struggle with science-domain inference procedures such as \textit{how to heat a substance} or \textit{how to grow a seed}, they also currently lack a fluency with commonsense skills such as \textit{navigating the environment} or \textit{storing liquids in containers}.  This underscores the need for models that can incorporate commonsense and science-domain knowledge, and integrate that declarative knowledge into actionable procedures to progress towards goals in the environment.
\vspace{4mm}

{\flushleft\textbf{Larger models are not necessarily better.}} While larger models generally perform better in question answering tasks \cite[\eg][]{2020t5}, here we observe that larger models do not always increase performance.  
Our best-performing model, the DRRN, has only 1.5 million parameters -- \textit{four orders of magnitude} less than the T5 models.  Both models also receive the same number of gradient updates ($10^6$) with respect to \textsc{ScienceWorld} training tasks---though the T5 models have the added benefit of pre-training both from science exam QA and a large number of expert demonstrations.\footnote{See the \textsc{Appendix} for additional experiments evaluating performance versus model size.} This underscores that how a model approaches modeling state spaces and action sequences may be more important than the scope of its pre-training. 
Online, interactive training enables models such as the DRRN and KG-A2C to perform tasks requiring long action sequences more efficiently in terms of both samples and parameters.

{\flushleft\textbf{Limitations of agents and environments.}}  While agents still find text environments challenging, it is important to recognize that even this modest performance is achieved through a number of simplifying properties.  For example, because agents frequently generate plausible but invalid actions, all but two agents benchmarked here depend on \textsc{ScienceWorld}'s valid action detection aid at test time, substantially simplifying their search problem in the action space.  Similarly, while \textsc{ScienceWorld} achieves a high environment fidelity for a text simulation, this is still tempered by pragmatic concerns, such as generating comparatively short descriptions of environments that can fit into the sequence lengths of most transformer models.  As such, even environments with complex physical, chemical, and biological processes underlying their simulations (such as \textsc{ScienceWorld}) still ultimately must limit the vividness of their descriptions, until these technical limitations in modelling can be surpassed.  
Hybrid environments \cite[\eg][]{shridhar2020alfworld} that concurrently model the same environment as both a high-fidelity 3D world and comparatively low-fidelity text-based simulation have shown that text environments can be used to provide useful task pre-training that can transfer back to the 3D environment with relatively low simulation compute cost. 

{\flushleft\textbf{Explanations as action sequences.}}  Explanations take on a variety of roles \cite{lombrozo2006structure,gilpin2018explaining}, from detailed human-readable descriptions of classification processes that describe \textit{how} a decision was made \cite[\eg][]{ribeiro2016should}, to higher-level appeals to scientific processes to explain \textit{why} an answer is correct \cite[\eg][]{jansen-etal-2018-worldtree,dalvi-etal-2021-explaining}.  Here, the action procedures generated by agents act as \textit{manner} explanations for how to solve a particular task -- but while they describe \textit{how} to accomplish something, they don't explain at a high-level \text{why} those actions accomplish the task.  For example, action sequences lack high-level goal information such as \textit{``melting a substance requires heating it, so first the agent needs to heat the substance with a heating device, like a stove.''}.  Similar to how cuing agents to answer contextual questions can help improve their task performance \cite{peng2021inherent}, cuing agents to generate these explanatory scaffolds may help future agents increase task performance, while structuring their action sequence explanations for better human interpretability.

\section{Conclusion}
Despite recent progress in both interactive text agents and scientific text processing via transformers, current models are unable to reason about fundamental science concepts in a grounded and reusable manner---calling into question how much they are actually understanding the tasks at hand.
To better measure such scientific reasoning abilities, we introduce \textsc{ScienceWorld}, an interactive text environment derived from an elementary school science curriculum---with tasks ranging from electrical conductivity to Mendelian genetics.

We evaluate three state-of-the-art reinforcement learning text game agents: DRRN, KG-A2C, and CALM; and further introduce two large-scale transformer-based agents inspired by recent advances such as Behavior Cloning and the Decision Transformer and trained for scientific reasoning in \textsc{ScienceWorld}.
While we find that overall performance on unseen tasks that require using science-domain knowledge is low across all agents, our results also suggest that agents that learn interactively in a grounded environment are more sample and parameter efficient than large language models that learn offline by reading text from static sources.
The best agent performance is still modest --- and on-par with medium-difficulty interactive fiction environments such as Zork --- highlighting the need for agents that can integrate declarative scientific and world knowledge with procedural action sequences in virtual environments. 



\section{Broader Impacts}
Interactive text environments can provide a faster and cheaper alternative to 3D environments to teach agents how to plan via sequential decision making. They allow better control over the level of abstraction desired to approach a task (i.e., \cmd{go to \token{kitchen}}, vs. \cmd{put \token{hand} on \token{door's knob}}, \cmd{turn \token{knob} clockwise}, \cmd{pull \token{door}}, \cmd{let go of the \token{knob}}, \cmd{walk through the \token{door}}). We believe making a plan in this abstract language space is simpler and more interpretable.

With respect to risks, we consider this current work as exploratory only. ScienceWorld is primarily intended for training agents to learn reasoning capabilities in the science domain, with limited immediate utility to human science students. Agents trained on ScienceWorld should not be used to provide advice for the real world. The environment in ScienceWorld has been made safer compared to the real world. 
For instance, the agent can't accidentally burn itself while boiling a substance on a campfire, and its actions should not be taken as demonstrations of safe procedures for students.


\section*{Acknowledgements}
This work supported in part by National Science Foundation (NSF) award \#1815948 to PJ, Google Cloud Compute, and the Allen Institute for AI.
The authors would also like to thank Liwei Jiang, Jack Hessel, and Oyvind Tafjord for their very timely technical assistance and advice, giving us the ability to train our larger, transformer-based agents. We'd also like to thank Michal Guerquin for technical assistance with the project website.

\bibliography{anthology,custom}

\clearpage
\appendix

\section{\textsc{ScienceWorld} Description}
\label{sec:appendix-scienceworld-description}
\textsc{ScienceWorld} is a simulation of the world abstracted through a complex interactive text environment with many objects (Sec. \ref{sec:Environment-and-Objects}), actions (Sec. \ref{sec:Action-Space}), and simulation engines (Sec. \ref{sec:Simulation-Engines}). The object-oriented (Sec. \ref{sec:Object-Model}) simulator is written in \textsc{Scala} and offers a \textsc{Python} interface to interact with it. \textsc{ScienceWorld}'s flexibility makes it simple to create new science-domain tasks (Sec. \ref{sec:specifictasks}) and to evaluate the correctness of agents' solutions (App. \ref{appsec:evaluation-protocol}).

\subsection{Object Model}
\label{sec:Object-Model}
Objects in \textsc{ScienceWorld} are represented using an object-oriented model and are implemented as classes.  \textsc{ScienceWorld} objects can be thought of as collections of sets of properties (\eg material properties, life properties, device properties, etc.).  All objects implement common functions, such as those that produce textual descriptions of the object, or that provide one or more possible referents for the object based on its current state (\eg the \token{water substance} in the solid state could generate the referents \token{ice}, \token{solid water}, and \token{substance}, each of which could be used by the agent to refer to that object in an action).  Similar to Z-machine games \cite{learningZIL1989}, objects are stored in an object tree 
representing the object's current container (its immediate parent object in the tree), and any objects it contains (child nodes in the tree). 

\subsection{Environment and Objects}
\label{sec:Environment-and-Objects}
\textsc{ScienceWorld} is composed of a map of 10 locations centered around a house theme (\token{kitchen}, \token{bathroom}, \token{workshop}, \token{art studio}, \token{greenhouse}, \token{outside}, \etc), as shown in Figure~\ref{fig:overview}.  While the rooms and how they interconnect is static, the environment is randomly populated with different combinations of relevant contextual items each time it is initialized -- for example, in one run, the living room may have a \token{bookcase} with three \token{books}.  In other runs, the bookcase may have different books, no books, or not be present in the environment.  This parametric variation discourages agents from memorizing the specific environment, and encourages robustness in task performance. 

The environment is populated with up to 195 specific types of objects (excluding variations of those objects that change names or task properties, \ie \token{red wires} and \token{black wires} belong to the same object type).  This includes 23 animals, 11 plants, 25 substances, 10 canonical liquid containers (like \token{tin cups} or \token{glass jars}), 13 electrical components (such as \token{light bulbs}, \token{motors}, \token{wires}, and \token{generators}), 16 devices (including a \token{stove}, \token{thermometer}, and \token{stopwatch}), and 15 common pieces of furniture.  To support these objects, the simulator includes a variety of other properties, including (for example) plant/animal life cycles, and 80 material properties (including water, glass, iron, and wood) that pure substances or physical objects (\eg \token{tables}) can be made from.

\subsection{Action Space}
\label{sec:Action-Space}
%
%

\begin{table}[t!]
\begin{center}
\footnotesize
\setlength{\tabcolsep}{3pt}
\begin{tabular}{ll} 
\toprule
\textbf{Action} &   \textbf{Description}\\
\midrule
\cmd{open/close \secondtoken{OBJ}}               & open/close a container            \\
\cmd{de/activate \secondtoken{OBJ}}              & activate/deactivate a device   \\
\cmd{connect \secondtoken{OBJ} to \secondtoken{OBJ}}              & connect electrical components  \\
\cmd{disconnect \secondtoken{OBJ}}              & disconnect electrical components  \\
\cmd{use \secondtoken{OBJ} [on \secondtoken{OBJ}]}   & use a device/item         \\
\cmd{look around}                          & describe the current room         \\
\cmd{look at \secondtoken{OBJ}}                  & describe an object in detail  \\
\cmd{look in \secondtoken{OBJ}}                  & describe a container's contents    \\
\cmd{read \secondtoken{OBJ}}                     & read a note or book       \\
\cmd{move \secondtoken{OBJ} to \secondtoken{OBJ}}      & move an object to a container \\
\cmd{pick up \secondtoken{OBJ}}                  & move an object to the inventory    \\
\cmd{put down \secondtoken{OBJ}}                 & drop an inventory item       \\
\cmd{pour \secondtoken{OBJ} into \secondtoken{OBJ}}    & pour a liquid into a container  \\
\cmd{dunk \secondtoken{OBJ} into \secondtoken{OBJ}}    & dunk a container into a liquid    \\
\cmd{mix \secondtoken{OBJ}}                      & chemically mix a container   \\
\cmd{go to \secondtoken{LOC}}                    & move to a new location    \\
\cmd{teleport to \secondtoken{LOC}} $^{*}$       & teleport to a specific room       \\
\cmd{eat \secondtoken{OBJ}}                      & eat a food                \\
\cmd{flush \secondtoken{OBJ}}                    & flush a toilet            \\
\cmd{focus on \secondtoken{OBJ}}                 & signal intent on a task object \\
\cmd{wait [\secondtoken{DURATION}]}              & take no action for some duration \\
\cmd{task}                                 & describe current task     \\
\cmd{inventory}                            & list agent's inventory    \\

\bottomrule
\end{tabular}
\caption{\footnotesize The 25 actions in the action space of \textsc{ScienceWorld}.  Actions can take up to two parameters, referencing objects the action should interact with.  $^{*}$ signifies that the \textit{teleport} action is only available to agents in a simplified  mode.
\label{tab:actionspace}}
\end{center}
\end{table}

The simulator implements 25 actions, shown in Table~\ref{tab:actionspace}, including generic actions common in interactive text environments (\eg opening a door, moving to a location), as well as science-domain specific actions (\eg connecting electrical components, chemically mixing items, pouring liquids).  Five actions take two arguments, 
16 take one argument, and four actions take zero arguments.  Given the approximately 200 possible objects (excluding parametric variations) in \textsc{ScienceWorld}, the action space can naively be estimated to be approximately 200,000 possible unique action possibilities at each step, though only a small subset of these would be meaningful. Similar to the Jericho framework, the simulator can provide valid action detection as an aid to agents (such as the DRRN) that require selecting their next action from a list of possible known-valid actions at a given step. 

\paragraph{Input Parser}
At each step, an input parser attempts to parse user or agent input into a single unique action.  Actions are specified as templates that can take on a variety of surface forms (\eg \cmd{move to} \token{LOCATION} or \cmd{go to} \token{LOCATION}), and that include placeholders for object referents.  At runtime, the parser examines all valid referents for visible objects from the agent's point of view, and if a given input string can produce more than one valid action, the parser will ask for clarification\footnote{For example, if the agent is in a room with two apples, one on a table and one in a bowl, the command \cmd{take \token{apple}} will cause the parser to ask the agent to clarify which apple they mean by selecting possible alternatives from a numbered list.}.

\subsection{Simulation Engines}
\label{sec:Simulation-Engines}
\textsc{ScienceWorld} supports actions commonly found in interactive text environments -- for example, objects can be moved or examined, foods can be eaten, and books can be read.  In addition, the environment contains a number of elementary science-domain specific processes that either occur automatically (\eg thermodynamics) or are coupled to actions (\eg using devices, mixing chemicals). 
Those simulation engines are:\footnote{To maintain tractability in implementation, simulation engines are implemented with a fidelity at the level of an elementary science curriculum.  Thermal transfer uses a simplified equation, biological changes happen in stages rather than gradually, only series instead of arbitrary electrical circuits are simulated (and without the concepts of resistance, inductance or other advanced topics), etc.} 

{\flushleft \textbf{Thermodynamics:}} All objects have temperatures and other thermal properties based on their materials.  All objects within a container are considered in thermal contact with each other, and transfer heat energy using a simplified conductive heat model.  The proportion of heat transferred between objects at each step is mediated by the object's \textit{thermal conduction coefficient}, allowing thermal conductors (like metal pots) and insulators (like ceramics) to be modelled.  Every material has phase transition points \textit{(\ie melting point, boiling point)} and \textit{combustion points} populated based on the best-known or approximate physical values for those materials.  Objects that move past these thresholds will change state of matter (\eg from a solid to a liquid), or begin a combustion process that ultimately ends in the object turning to ash unless its fire is put out.  Convective heat transfer is also modelled in the form of heat sources (\eg \token{oven}, \token{stove}) and heat sinks (\eg \token{fridge}, \token{freezer}) that transfer heat energy to or from objects.  Rooms also transfer ambient heat energy to/from the objects they contain.

{\flushleft \textbf{Electricity:}} The simulator models simple series electrical circuits, where electrically-powered devices (\eg \token{light bulb}, \token{motor}) can be powered by being connected to electrical energy sources (\eg \token{battery}, \token{solar panel}) through electrical conductors (nominally, \token{wires}).  Polarized and unpolarized components are modelled, with each object having exactly two terminals (either an anode and cathode for polarized components, or terminals 1 and 2 for unpolarized components).  Connection happens through explicit terminal-to-terminal connection actions (\eg \cmd{connect \token{battery anode} to \token{blue wire terminal 1}}).  Every non-electrical object in \textsc{ScienceWorld} has virtual unpolarized terminals, allowing circuits to be build with valid electrical conductors (\eg using a \token{metal fork} in place of a \token{wire}), and for the agent to build circuits that test conductivity by (for example) observing if a light bulb illuminates when a plastic versus metal fork is used in the circuit.

{\flushleft \textbf{Devices:}} Many objects are also considered devices, that can be activated or deactivated by the agent (\eg \token{stove}, \token{electrical switch}), or may have environment-specific conditions to being activated (\eg a \token{light bulb} will only activate if it is properly electrically connected; a \token{solar panel} will only produce power if it is outside). Objects can also be used with other objects in specific contexts (\eg a \token{thermometer}, to measure an object's temperature; a \token{shovel}, to dig soil from the ground).  

{\flushleft \textbf{Chemistry:}} A subset of specific chemical reactions are modelled, where mixing a set of substances together in a container will produce a resultant substance (\eg \token{salt} and \token{water} mix to produce \token{salt water}).  Common chemical reactions described in elementary science questions (\eg water reactions, rust, food reactions, paint mixing) are modelled.

{\flushleft \textbf{Life Stages:}} Living things (plants and animals) progress through life stages (\eg seed, seedling, juvenile plant, adult plant, reproducing plant, dead plant).  Progression through life stages happens over time by continuing to meet the needs of that living thing (\eg \token{water}, \token{soil}).  If the needs are not met (\eg a plant is not watered, is removed from soil, or becomes too hot), then it dies.

{\flushleft \textbf{Reproduction and Genetics:}} Living things can have genes that express traits (\eg flower colour, seed shape, leaf size).  Genes are inherited from the alleles of both parents, and genotype is determined at the time of reproduction using a Punnett square.  Phenotype (expressed, visible traits) are determined based on which genes are dominant versus recessive.  Currently, traits are only populated for selected plants to reproduce Mendelian-genetics experiments.  Plants reproduce by exchanging pollen (containing their genes) between flowers, typically by a pollinator (such as a \token{bee}).  Pollinated flowers eventually wilt and turn into fruits containing seeds of genetic descendants. 

{\flushleft \textbf{Friction (Inclined Plane):}} Forces are a significant part of an elementary science curriculum, but difficult to incorporate without 2D or 3D simulation.  \textsc{ScienceWorld} models the forces of gravity and friction in the specific context of 1-dimensional inclined plane experiments.  Objects placed at the top of an inclined plane will slide down the plane at a speed proportional to the plane's angle, and the friction coefficient of its surface material.  The position is described to the agent (\eg \textit{``an inclined plant with a block 60\% of the way down the plane''}), allowing experiments to determine either the relative angle or friction coefficients of different inclined planes based on the speed the object moves down a given plane. 

{\flushleft \textbf{Containers:}} Containers can be always open (\eg a \token{metal pot}) or closeable (\eg a \token{cupboard}).  Objects contained inside containers are not visible until the container is open.  Some effects spread beyond a container -- for example, a wooden cupboard with a hot object inside may combust, causing other objects in the kitchen to also increase temperature.


\section{Tasks and Competencies}
To support our goal of generating a diverse set of tasks, we identified a candidate set of 10 broad science exam topics from the list of 400 fine-grained science curriculum topics of \citet{xu-etal-2020-multi}.  Topics were chosen that would be amenable to text-based simulation, and that did not have critical fine-grained spatial reasoning requirements.  These topics and tasks are described in Section~\ref{sec:specifictasks}.


{\flushleft \textbf{Subtasks and Masked Objects:}} Each of the 10 broad curriculum topics is further subdivided into between 2 and 4 specific subtasks that test specific reasoning capacities (\eg \textit{melting, boiling, and freezing} subtasks for the change-of-state task), or ask the agent to perform the same task but with names of critical task objects masked.  Some tasks are possible to partially solve by looking up critical task information (\eg knowing that \token{white flowers} are a dominant trait of pea plants for the Mendelian genetics task). We include two versions of tasks, one with using masked names (\eg growing \token{Unknown Plant B} instead of a \token{Pea Plant}) while simultaneously randomly generating the properties of those objects to provide an instrument to measure when agents are solving tasks by performing the experimental procedure, and when they are directly looking up answers. 


{\flushleft \textbf{Task Formats:}} Task goals are structured with the broad goal of both (a) accomplishing a task, and (b) doing so \textit{intentionally}.  Tasks typically include preliminary subgoals where the agent must signal their intent to perform the task on a specific object by first  \textit{``focusing''} on the object they intend to perform the task with (\eg for a \textit{boiling} task, focusing on water they intend to boil) before they perform the task. 

Tasks take on two main forms: \textit{Perform a task:} the agent must directly perform a task, that produces some measurable change in the environment (\eg growing a fruit through pollination) that can be directly measured as an end-state.  \textit{Forced-choice:} The agent must perform a task that requires making an inference (\eg whether an object is an electrical conductor or insulator), and provide their answer by placing the task object in a specific container (\ie an \textit{``answer box''}) if the object is conductive, and a different container if it is an insulator. 


{\flushleft\textbf{Task Variations:}} To prevent overfitting and encourage generalization, each subtask contains between 10 and 1400 parametric variations of that subtask (with 7200 total variations across all 30 subtasks).  Variations change critical task objects (\eg the specific substance to be melted), the agent's starting location in the environment, as well as randomly vary the contents of the environment itself (\eg whether the living room contains a bookshelf, or a painting, or both).

{\flushleft\textbf{Task Simplifications:}}
Agents find different competencies that \textsc{ScienceWorld} tests to be challenging.  Tasks can be made easier by enabling any of 5 environment simplifications (or choosing ``easy'' mode, which enables all the simplifications).  Examples of simplifications include a \textit{teleport} action that lets agents instantly move to any location, and having all containers open by default.

\subsection{Scoring and Evaluation Protocol}
\label{appsec:evaluation-protocol}
{\flushleft\textbf{Goals and Reward Shaping:}} Each subtask contains a small number of method-agnostic required goals to be met (such as \textit{focusing on the substance to melt}, and \textit{causing that substance's state of matter to change from a solid to a liquid} for the melting task).  In addition, to make rewards less sparse for agents learning these tasks, each task includes between 2 and 15 optional subgoals (such as \textit{turning on the stove}, or \textit{the substance increasing in temperature by 10C}) that help nudge agents in the direction of canonical solutions, if desired.  Meeting required and optional subgoals increases the agent's score on a given subtask.  Scores for all tasks are normalized to between 0 and 1.

\subsection{Specific Tasks}
\label{sec:specifictasks}
{\flushleft \textbf{Changes of State:}} The agent must find a named substance (\eg \token{ice}), and change the state of matter of that substance (solid, liquid, gas) using the heating and cooling devices (\eg \token{stove}, \token{freezer}) available in the environment.  Subtasks require specific phase changes (melting, boiling, freezing, or the agent's choice).  Variations change the substance, and ablate common devices (\eg the stove becomes disabled) so that the agent must find alternate methods of heating or cooling. 

{\flushleft \textbf{Measurement Instrument:}} The agent must find a \token{thermometer} and use it to measure the temperature of a named object.  In two additional subtasks, the agent must use the thermometer to measure the melting point of a named substance by heating it and continually monitoring the temperature.  Answers are modelled as a forced-choice task, where the agent is given a predetermined temperature threshold at the start of the task (\eg $50^{\circ}C$), and must focus on one answer box if the melting point is above the threshold, and the other answer box if the melting point is below the threshold.  Variations change the substance to be measured, and the preset temperature threshold.

{\flushleft \textbf{Electrical Circuits:}} The agent must build a working series electrical circuit by connecting various electrical components including power sources (\eg \token{battery}, \token{wind mill}, \token{gas generator}), different coloured \token{wires}, and active components (\eg \token{lights}, \token{motors}).  Subtasks include (a) powering a named component, (b) powering using renewable versus nonrenewable energy, and (c) measuring whether named or unknown substances are electrically conductive.  Variations change available components, colours of wire, and specific task objects required to be used in the circuit.

{\flushleft \textbf{Classification:}} In four subtasks, the agent must find an object in the environment belonging to a specific category (living things, non-living things, plants, or animals), and place it in an answer box.  Variations change the location and description of the answer box.

{\flushleft \textbf{Growing plants:}} The agent must grow a named plant (\eg a \token{peach tree}) from seed.  To do this, they must place the \token{seed} and \token{soil} in a \token{flower pot}, and provide regular water as the plant progresses through life stages into adulthood. 
Failure to water appropriately causes the plant to perish.  In a subtask, the agent must grow a fruit by growing several plants, 
then releasing pollinators (\ie \token{bees}) that cross-pollinate flowers on the plants, which will eventually produce fruits.  Variations change seed type, and soil location (either provided in the pot, provided in the room, or must be gathered outside using a shovel).

{\flushleft \textbf{Chemistry:}} The agent must create a specific substance by mixing two or more input substances in a container.  In the generic subtask, a recipe document that can be read by the agent is provided somewhere in the environment.  In two paint-themed subtasks, the agent is given primary colours of paint (red, green, yellow), and must mix secondary (\eg orange) or tertiary (\eg orange-yellow) colours through several steps.  Variations change the required output substance.

{\flushleft \textbf{Life Spans:}} In three task variations, the agent must find 3 animals in the environment, then select either the shortest-lived (\eg \token{bee}), longest-lived (\eg \token{turtle}), or shortest-then-longest lived (bee-then-turtle).  Variations change which animals are populated in the environment from a list of long, medium, and short-lived animals.

{\flushleft \textbf{Life Stages:}} The agent must find a named plant or animal, and focus on its life stages, from earliest (\eg seed) to latest (\eg reproducing adult plant).  Variations change the plant or animal involved. 

{\flushleft \textbf{Forces:}} The agent must use inclined planes and masses (\eg a \token{block}) for experiments about forces.  In one subtask the agent is given two planes, and must determine which has the steepest or shallowest angle based on the time the block takes to move down the plane.  In two other subtasks, the agent must find which of two planes has a surface of highest or least friction, from either named (\eg plastic, steel) or unknown surfaces.  The agent can measure time internally (in terms of number of steps for a block to fall), or measure this explicitly with a provided stopwatch.  Variations change inclined plane angles (first task) or surface material types (remaining tasks). 

{\flushleft \textbf{Mendelian Genetics:}} The agent must determine whether a named trait (\eg white flower colour) is a dominant or recessive trait in a plant.  Two seeds are provided (one with the trait as dominant, one recessive), and the agent must grow two generations of plants and count how often it observes a given trait in successive generations to determine whether it is dominant or recessive.  Subtasks change whether the plant is known (the \token{pea plant}, as in Mendel's experiments) or a randomly generated plant, while variations change the trait under investigation.

\subsection{Commonsense Competencies}
In addition to science-domain competencies, the agent must demonstrate fluency with commonsense knowledge and procedures to complete tasks successfully.  Agents must know the locations of common objects (\eg \token{water} comes from a \token{sink}, \token{orange juice} is typically found in a \token{fridge}), the affordances of common objects (a \token{sink} can be turned on to create \token{water}, a \token{cup} can be used to carry a liquid), navigation (a world is made of discrete rooms that can be traversed through doors), containers need to be opened to observe or use their contents, and so forth.


\subsection{Scoring and Evaluation Protocol}
\label{appsec:evaluation-protocol}
{\flushleft\textbf{Goals and Reward Shaping:}} Each subtask contains a small number of method-agnostic required goals to be met (such as \textit{focusing on the substance to melt}, and \textit{causing that substance's state of matter to change from a solid to a liquid} for the melting task).  In addition, to make rewards less sparse for agents learning these tasks, each task includes between 2 and 15 optional subgoals (such as \textit{turning on the stove}, or \textit{the substance increasing in temperature by 10C}) that help nudge agents in the direction of canonical solutions, if desired.  Meeting required and optional subgoals increases the agent's score on a given subtask.  Scores for all tasks are normalized to between 0 and 1. 

{\flushleft\textbf{Train, Development, Test sets:}} For a given subtask, variations are split into 50\% training, 25\% development, and 25\% test sets.  Variations are sorted such that critical unseen variations (\eg substances, animals, or plants unseen during training) are found in development and test sets. 

\subsection{Task Simplifications}
Agents find different competencies that \textsc{ScienceWorld} tests to be challenging.  Tasks can be made easier by enabling any of 5 environment simplifications (or choosing ``easy'' mode, which enables all the simplifications).  Examples of simplifications include a \textit{teleport} action that lets agents instantly move to any location; self-watering flowering flower pots that mean plants do not have to be frequently watered; and having all containers open by default.

\section{Experiment Details}
\label{sec:appendixDetails}
\subsection{Reinforcement Learning Models}
For each reinforcement learning model, we ran 8 environment threads at 100k steps per thread. Training episodes reset after meeting an end state (success or failure), or after reaching 100 steps. For KG-A2C and CALM, the training episode will also reset if stuck by invalid actions for 100 steps (invalid actions are not counted by the environment). We did evaluation on the test variations every 1000 steps per environment thread. We randomly chose 10 test variations and run 1 episode of testing for each chosen variation during each evaluation period and reported the average score of the 10\% test steps scores. 

{\flushleft\textbf{DRRN:}} We use the DRRN architecture from \url{https://github.com/microsoft/tdqn}, with embedding size and hidden size set as 128. The learning rate we use to train DRRN is 0.0001. The memory size is 100k, and priority fraction is 0.50.
The tokenizer for the input text is a uni-gram subword tokenizer model adapted from \citet{kudo-2018-subword}.

{\flushleft\textbf{KG-A2C:}} We make two major changes to the original KG-A2C model to function with \textsc{ScienceWorld}. (1) We replace the OpenIE knowledge graph extractor with a heuristic extractor.  The heuristic extractor uses regular expressions to parse the text of the ``look around'' and ``agent inventory'' information into \textit{(subject, relation, object)} triples.  This heuristic functionally extracts the ground truth knowledge graph representation of the observable environment for the agent. (2) We change the KG-A2C agent to generate object \textit{types} instead of references to specific objects. After selecting the action template and object type fillers that the agent has the highest confidence in, we ground those object types (e.g. apple) with specific object referents in the agent's current visible environment.  If more than one referent meets that type, one is chosen at random. 
The learning rate we use to train the KG-A2C agent is 0.003 and the tokenizer used is the same as the DRRN agent.

{\flushleft\textbf{CALM-GPT2:}} Following the original CALM paper \cite{yao-etal-2020-keep}, we use a 12-layer, 768-hidden, 12-head GPT-2 model. We use the default pre-trained weight offered by the Huggingface Transformers library \cite{wolf-etal-2020-transformers}. We fine-tune this GPT-2 model on complete action sequences generated by the oracle agents. The GPT-2 input prompt is formed as ``[CLS] $d$ [SEP] $o_t$ [SEP] $o_t^\textrm{look}$ [SEP] $o_t^\textrm{inv}$ [SEP] $o_{t-1}$ [SEP] $a_{t-1}$ [SEP]'' and targets to predict $a_t$, where $d$ stands for the task description and $o_t$, $o_t^\textrm{look}$, $o_t^\textrm{inv}$, and $a_t$ stand for the observation (excluding the look around and inventory information), the output of a ``look around'' action at the agent's current location, the agent inventory, and the action at time step $t$. We use beam search for generation, generating 16 beams representing candidate actions for the agent to select from. 
We set the diversity penalty to 50.0 to encourage the GPT-2 model to generate different actions. For the GPT-2 training we use a batch size of 12 and train for 20 epochs with a learning rate of 0.00002. The learning rate we use to train the CALM agent is 0.0001 and the input tokenizer is the same as that used in the original GPT-2 paper.

Due to the high computation cost of the CALM model, and modest overall performance, performance for each task is the average of only 3 random seeds instead of the 5 used for training the DRRN and KGA2C models.  During development, a small error was found in the prompt.  Pilot experiments suggested this resulted in a negligible ($\pm 0.01$) change in performance, so the full experiments (requiring up to 6000 GPU hours) were not regenerated.

\begin{figure*}[t!]
\centering
\begin{subfigure}{.24\textwidth}
  \includegraphics[width=\textwidth]{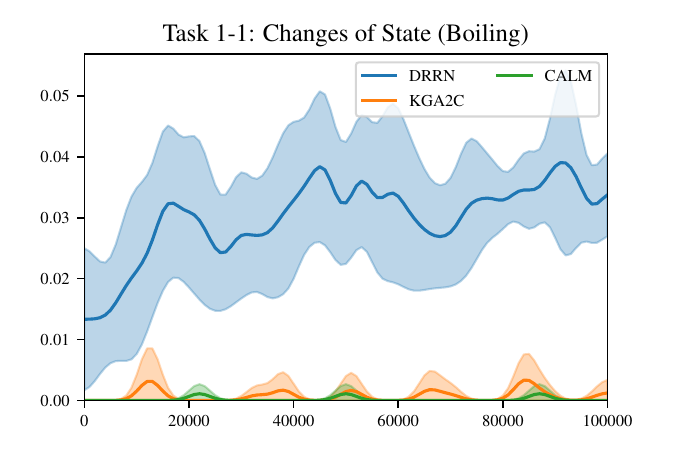}
\end{subfigure} 
\begin{subfigure}{.24\textwidth}
  \includegraphics[width=\textwidth]{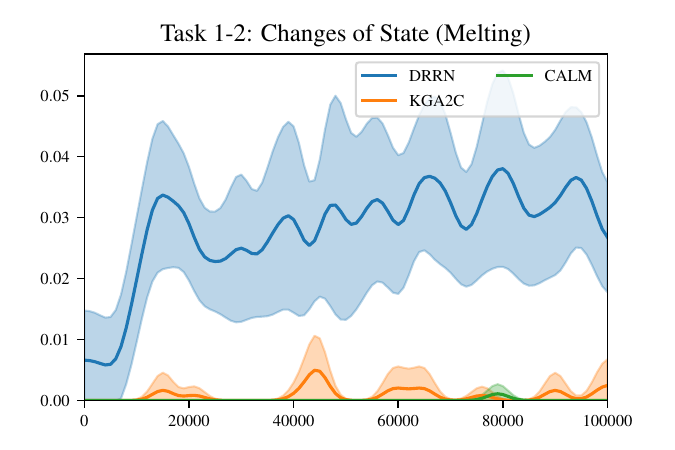}
\end{subfigure}
\begin{subfigure}{.24\textwidth}
  \includegraphics[width=\textwidth]{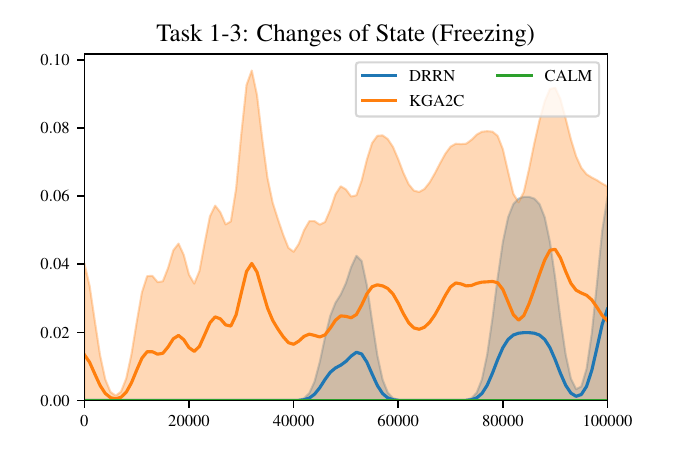}
\end{subfigure} 
\begin{subfigure}{.24\textwidth}
  \includegraphics[width=\textwidth]{plots-oct2022/task1-4.pdf}
\end{subfigure}
\medskip 
\begin{subfigure}{.24\textwidth}
  \includegraphics[width=\textwidth]{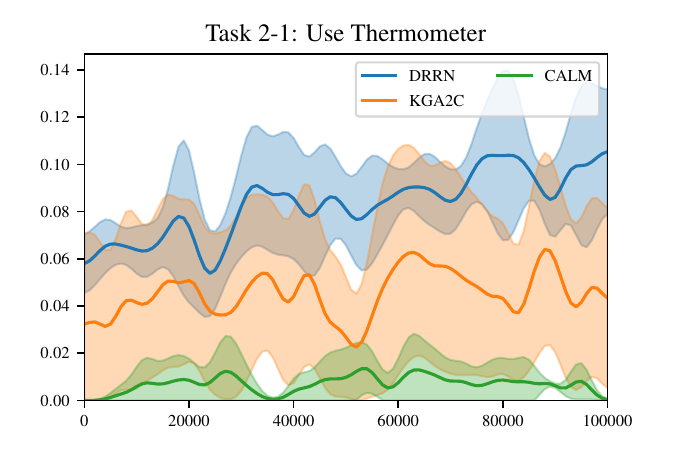}
\end{subfigure}
\begin{subfigure}{.24\textwidth}
  \includegraphics[width=\textwidth]{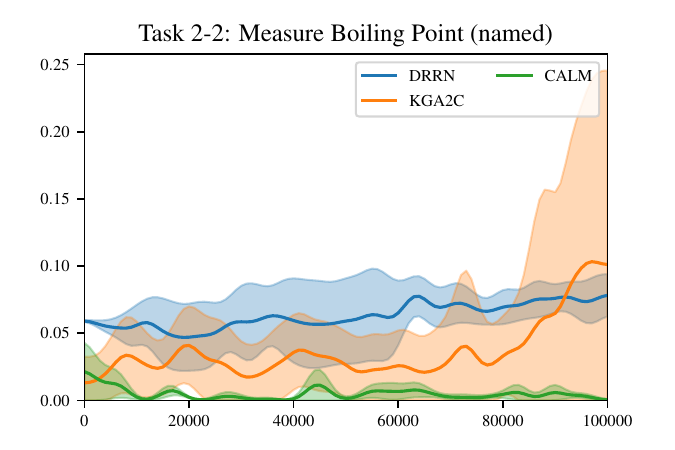}
\end{subfigure}
\begin{subfigure}{.24\textwidth}
  \includegraphics[width=\textwidth]{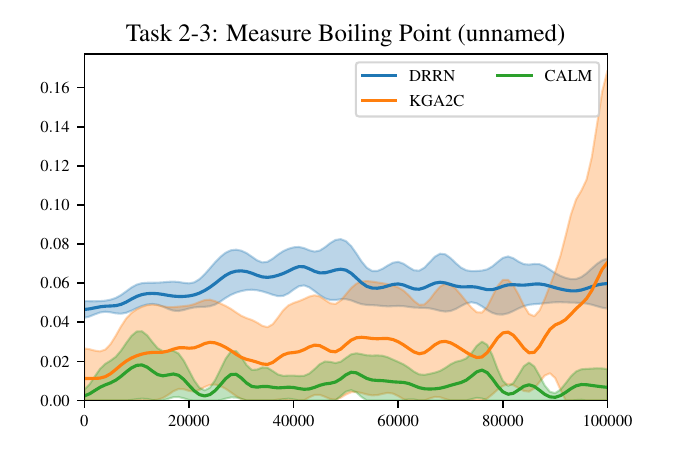}
\end{subfigure}
\begin{subfigure}{.24\textwidth}
  \includegraphics[width=\textwidth]{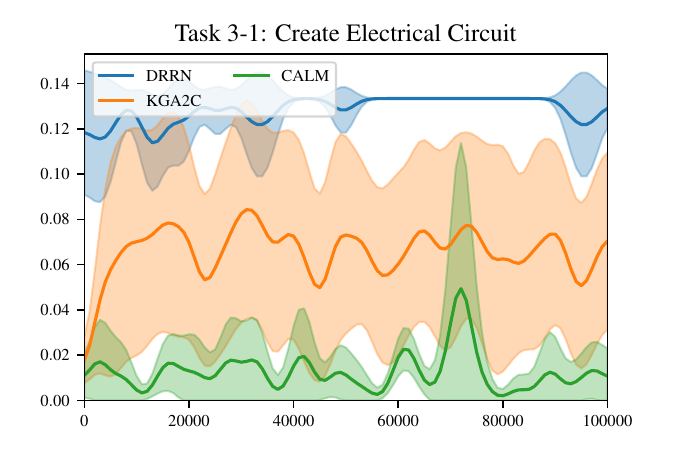}
\end{subfigure}
\medskip 
\begin{subfigure}{.24\textwidth}
  \includegraphics[width=\textwidth]{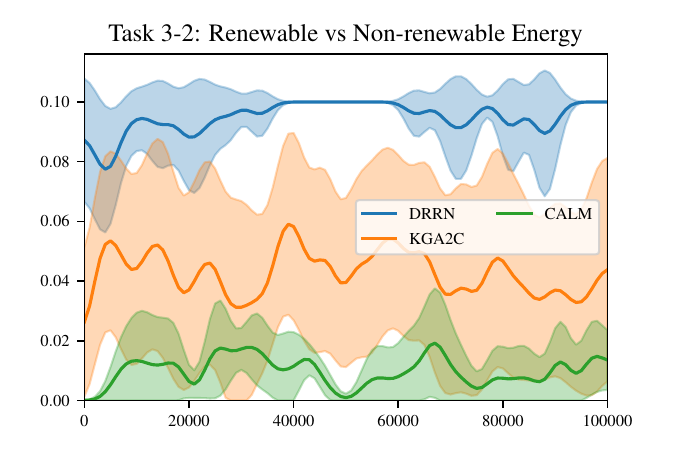}
\end{subfigure}
\begin{subfigure}{.24\textwidth}
  \includegraphics[width=\textwidth]{plots-oct2022/task3-3.pdf}
\end{subfigure}
\begin{subfigure}{.24\textwidth}
  \includegraphics[width=\textwidth]{plots-oct2022/task3-4.pdf}
\end{subfigure}
\begin{subfigure}{.24\textwidth}
  \includegraphics[width=\textwidth]{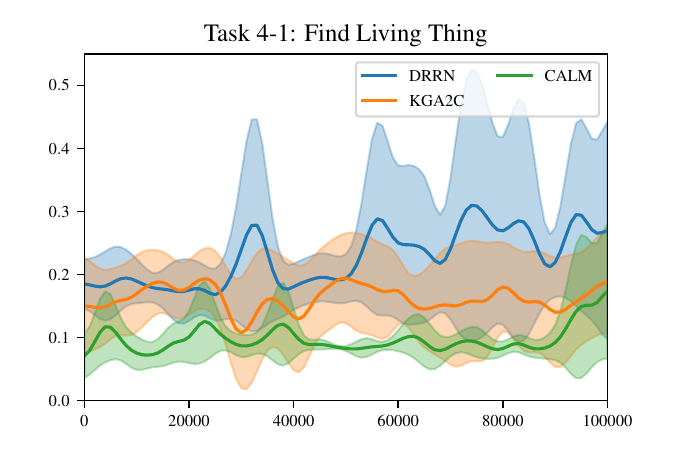}
\end{subfigure}
\medskip 
\begin{subfigure}{.24\textwidth}
  \includegraphics[width=\textwidth]{plots-oct2022/task4-2.pdf}
\end{subfigure}
\begin{subfigure}{.24\textwidth}
  \includegraphics[width=\textwidth]{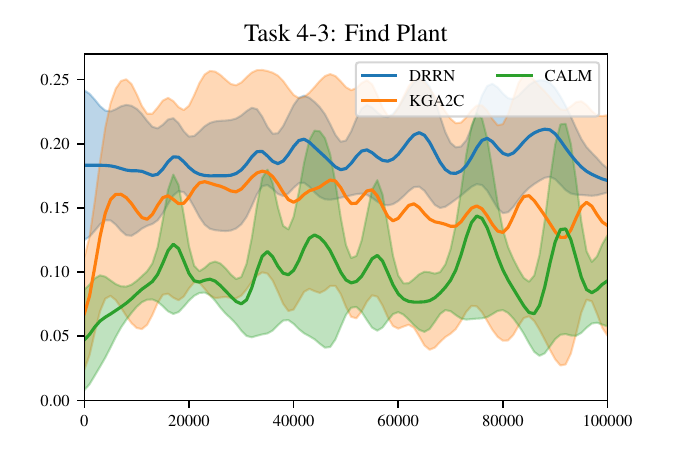}
\end{subfigure}
\begin{subfigure}{.24\textwidth}
  \includegraphics[width=\textwidth]{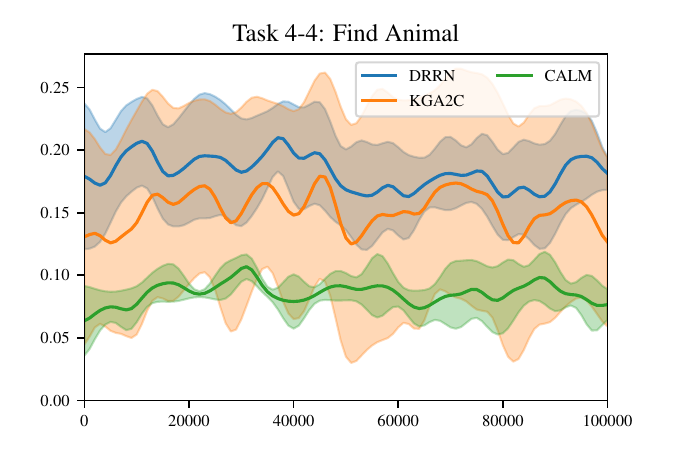}
\end{subfigure}
\begin{subfigure}{.24\textwidth}
  \includegraphics[width=\textwidth]{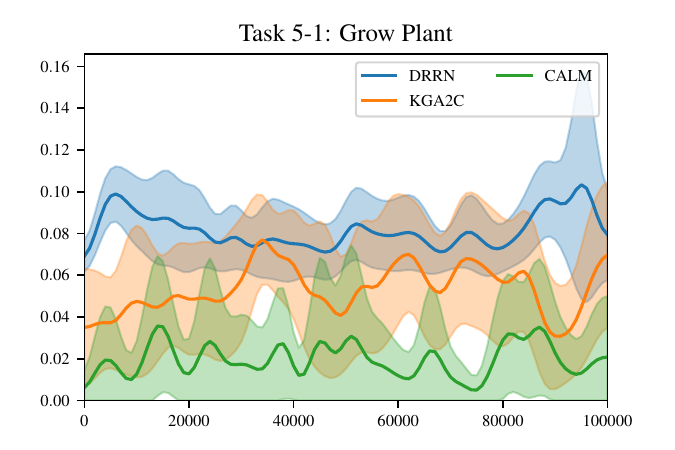}
\end{subfigure}
\medskip 
\begin{subfigure}{.24\textwidth}
  \includegraphics[width=\textwidth]{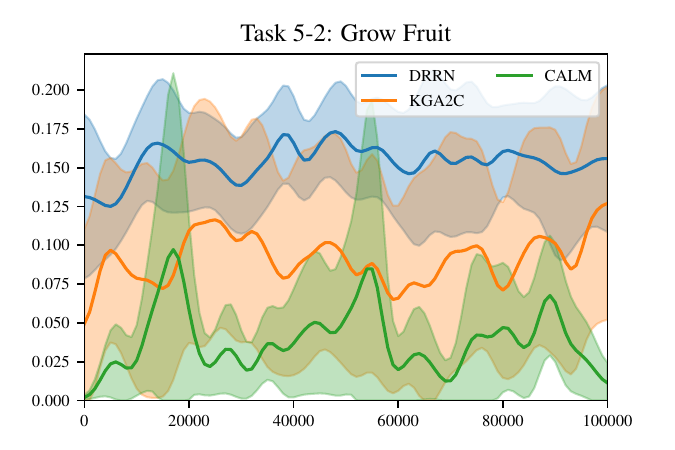}
\end{subfigure}
\begin{subfigure}{.24\textwidth}
  \includegraphics[width=\textwidth]{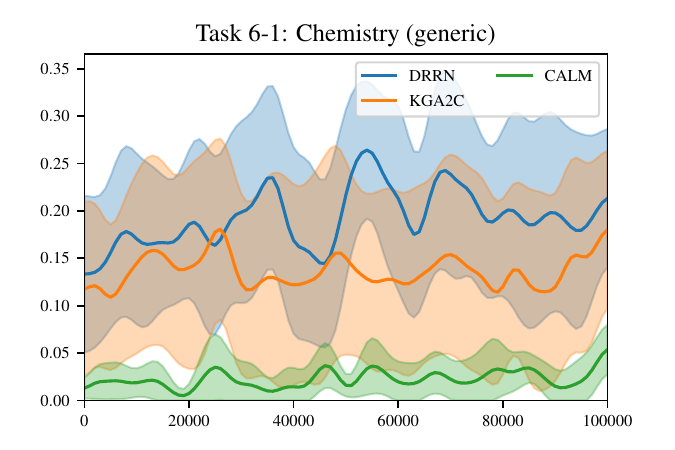}
\end{subfigure}
\begin{subfigure}{.24\textwidth}
  \includegraphics[width=\textwidth]{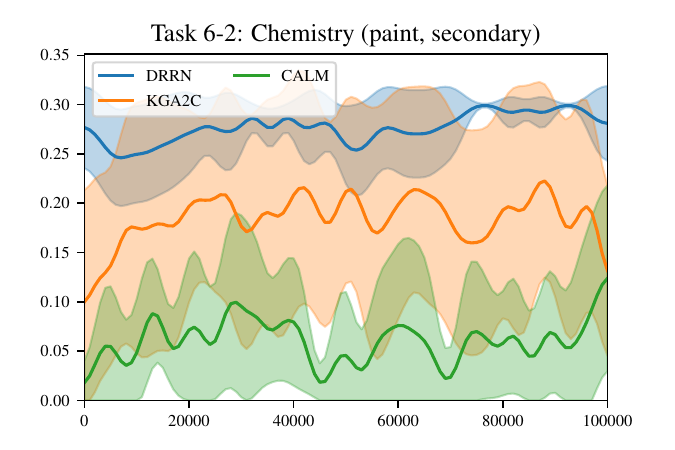}
\end{subfigure}
\begin{subfigure}{.24\textwidth}
  \includegraphics[width=\textwidth]{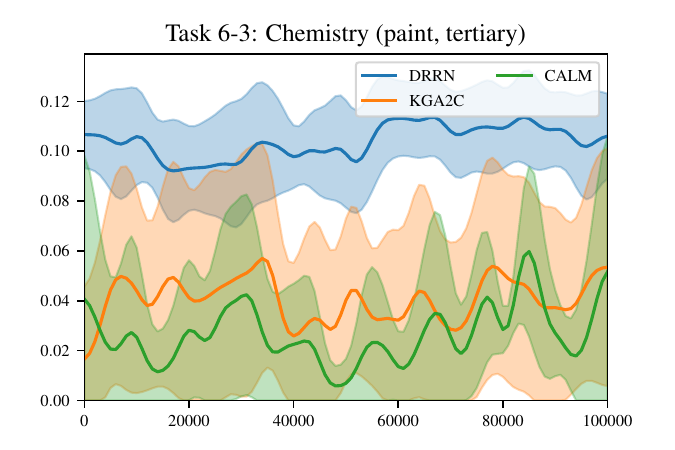}
\end{subfigure}
\medskip 
\begin{subfigure}{.24\textwidth}
  \includegraphics[width=\textwidth]{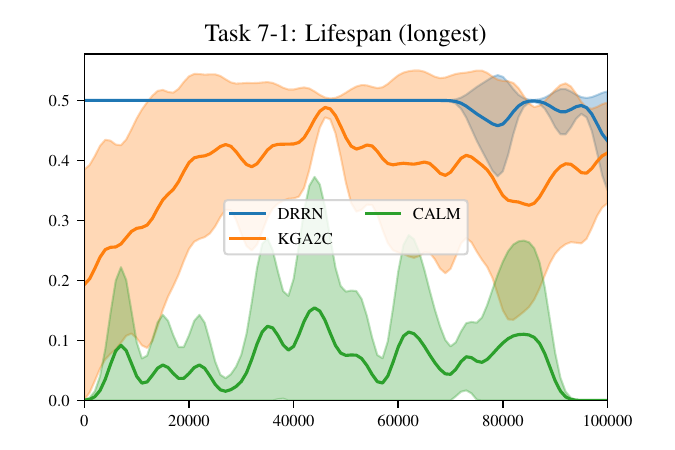}
\end{subfigure}
\begin{subfigure}{.24\textwidth}
  \includegraphics[width=\textwidth]{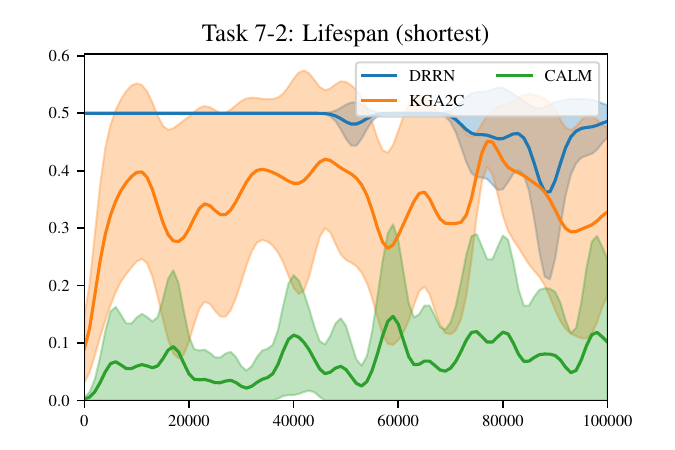}
\end{subfigure}
\begin{subfigure}{.24\textwidth}
  \includegraphics[width=\textwidth]{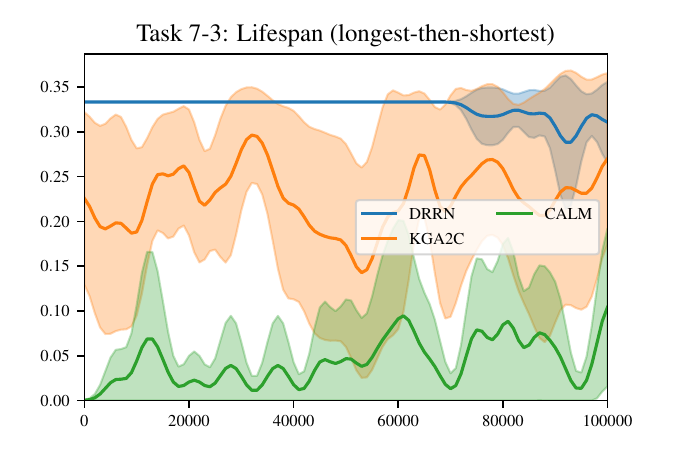}
\end{subfigure}
\begin{subfigure}{.24\textwidth}
  \includegraphics[width=\textwidth]{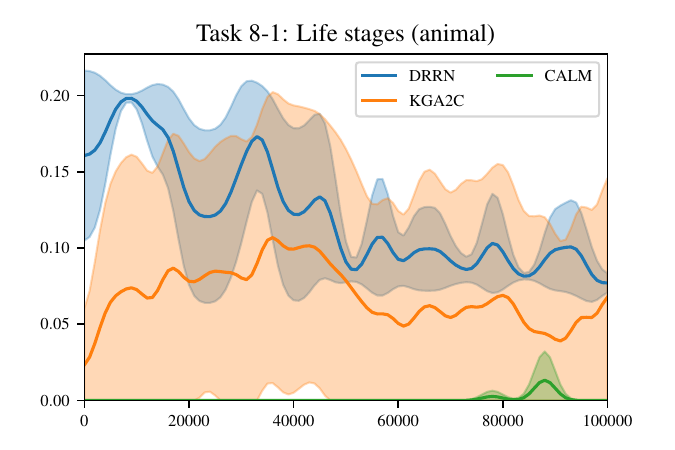}
\end{subfigure}
\medskip 
\begin{subfigure}{.24\textwidth}
  \includegraphics[width=\textwidth]{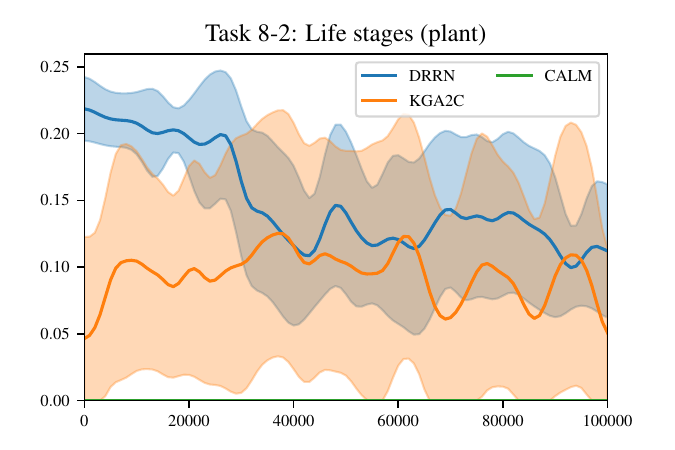}
\end{subfigure}
\begin{subfigure}{.24\textwidth}
  \includegraphics[width=\textwidth]{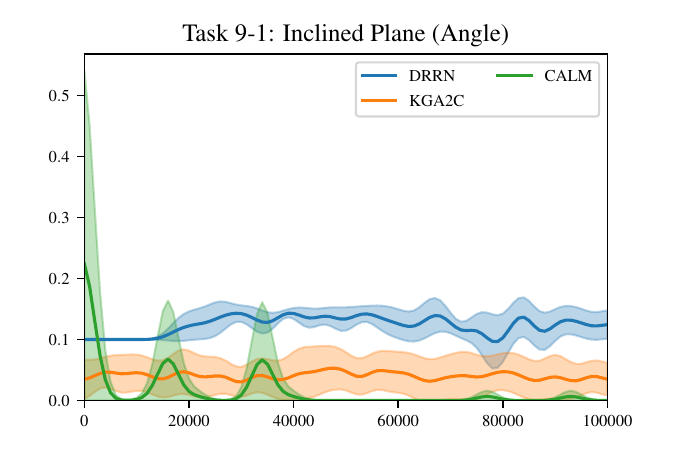}
\end{subfigure}
\begin{subfigure}{.24\textwidth}
  \includegraphics[width=\textwidth]{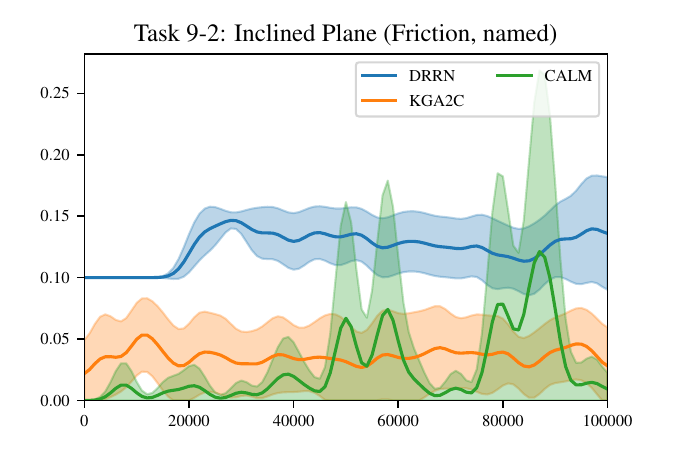}
\end{subfigure}
\begin{subfigure}{.24\textwidth}
  \includegraphics[width=\textwidth]{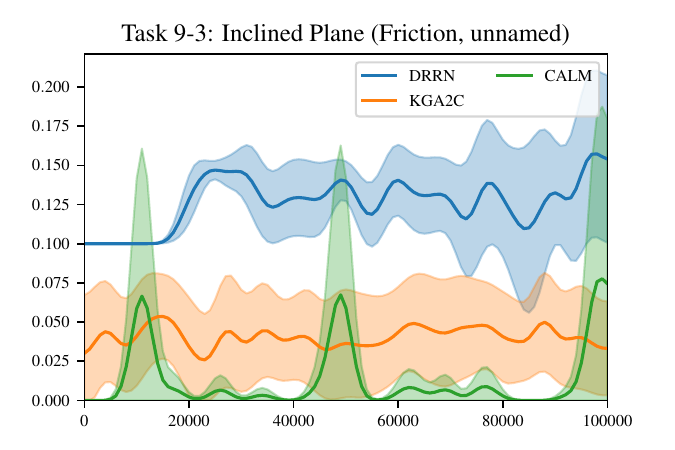}
\end{subfigure}
\medskip 
\begin{subfigure}{.24\textwidth}
\end{subfigure}
\begin{subfigure}{.24\textwidth}
  \includegraphics[width=\textwidth]{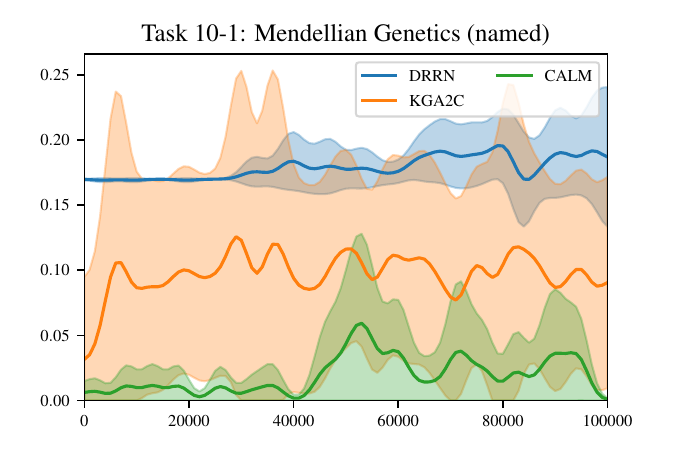}
\end{subfigure}
\begin{subfigure}{.24\textwidth}
  \includegraphics[width=\textwidth]{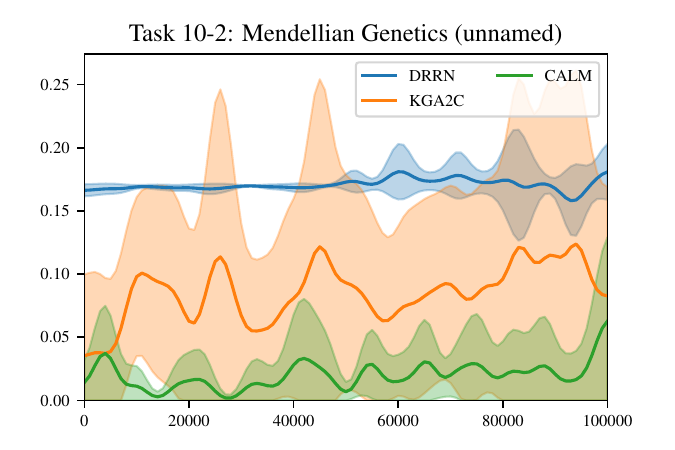}
\end{subfigure}
\begin{subfigure}{.24\textwidth}
\end{subfigure}

%

\caption{Episode reward curves for the DRRN, KGA2C, and CALM models on the unseen test set as a function of the number of training environment interactions (steps).}
\label{fig:rlcurves}
\end{figure*}

{\flushleft\textbf{Episode reward curves:}} Episode reward curves for each model across all 30 subtasks in \textsc{ScienceWorld} are shown in Figure~\ref{fig:rlcurves}.

\subsection{Behavior Cloning and Text Decision Transformer}
The T5 models used to train both of these models are initialized with weights and tokenizers from the trained Macaw-11b model released at \url{https://github.com/allenai/macaw}.
They are trained on a v3-32 TPU pod with a batch size of 16 and 32-way model parallelism for 100k gradient update steps.

At inference time, we use the model to generate actions given the observation of current and previous step. We use beam search with a beam size 16 to get the top 16 generations. We set the diversity penalty to 50.0 to encourage diversity in generation. For each subtask, we run one episode on each of its test variations and report the average score. We do not update weights of the T5 model during evaluation.

\subsection{Resources}
\begin{table}[h!]
\begin{center}
\footnotesize
\setlength{\tabcolsep}{3pt}
\begin{tabular}{lcc} 
\toprule
\textbf{Model} & \textbf{GPU} & \textbf{Runtime}\\
\midrule
\textit{Online RL Models}\\
~~DRRN             & 4gb & 12h    \\
~~KG-A2C           & 16gb & 20h   \\
~~CALM-GPT2        & 16gb & 40h   \\
\midrule
\textit{Offline Transformer Models}\\
~~Model Pre-training    & v3-32 TPU Pod   & 60h\\
~~BC-T5            & 3x 48gb & 2h \\
~~TDT-T5           & 3x 48gb & 2h \\
\bottomrule
\end{tabular}
\caption{\footnotesize Approximate computational resources per model.  Online RL and Offline Transformer model reflect runtimes for one full run of one subtask at one seed. Runtime estimates should be multiplied by the number of tasks (30) and number of random seeds (5) to obtain full runtime estimates.  All GPU-based runs were completed on P100, V100. or A6000 GPUs. \label{tab:experiment_resources}}
\end{center}
\end{table}



\clearpage

\subsection{Impact of Model Size and Pre-training Methodology on Performance}
\begin{table}[!h]
\centering
\footnotesize
\vspace{6mm}

\begin{tabular}{lcc}
\toprule
                 & \textbf{Average}         & \textbf{Model} \\
\textbf{Model}   & \textbf{Performance}     & \textbf{Parameters} \\
\midrule
DRRN    &   0.17      &   1.5M  \\
KGA2C    &   0.11      &   5.5M  \\
CALM    &  0.05      &   131M\textsuperscript{*}  \\
\midrule
\textit{Behavior Cloned} \\
~~~T5-Large          &   0.15  &   770M  \\
~~~Macaw-Large       &   0.17  &   770M  \\
~~~Macaw-11B         &   0.08  &   11B  \\

\midrule
\textit{Decision Transformer} & ~ & ~ \\
~~~T5-Large          &   0.13  &   770M  \\
~~~Macaw-Large       &   0.15  &   770M  \\
~~~Macaw-11B         &   0.08  &   11B  \\

\bottomrule
\end{tabular}
\caption{\footnotesize Average model performance versus model parameter size across all tasks and random seeds.  For \textsc{ScienceWorld} tasks, larger models do not necessarily perform better, and in some cases appear to show inverse scaling. \footnotesize $^{*}$ signifies that the value of 131M parameters includes the number of the parameters of the pre-trained GPT-2 action generator model. Only 6.9 million policy parameters are updated in RL training.}
\label{tab:inverse-scaling} 
\end{table}

To examine the effect of model size on behavior cloned and decision transformer model performance, we ran two model sizes for the T5 models, shown in Table~\ref{tab:inverse-scaling}.  We first observe that pre-training specifically for scientific question answering on a curated dataset (Macaw) outperforms T5 general language model pre-training. Further, we note that T5-Large and Macaw-Large, with 14 times fewer parameters (770m each), out-perform the larger 11 billion parameter models by approximately a factor of two.  These results suggest that while \textsc{ScienceWorld} can benefit from external scientific knowledge, it may also present an inverse scaling problem\footnote{\url{https://github.com/inverse-scaling/prize}}, where increasing model size can sometimes decrease overall task performance.  However, these results are only suggestive of an inverse scaling problem rather than a concrete demonstration. Due to the high cost of training and inference for these models, we can't currently rule out that these differences may be due to hyperparameter differences.  We leave this verification as future work.

\subsection{Attribution}
Graphical visualization makes use of RPG game assets developed by @Noiracide.

\end{document}